%% file: main.tex
\documentclass[acmtog,preprint]{acmart}
\acmSubmissionID{184}

\usepackage{booktabs} 

\usepackage[ruled]{algorithm2e} 
\usepackage{multirow}

\SetAlFnt{\small}
\SetAlCapFnt{\small}
\SetAlCapNameFnt{\small}
\SetAlCapHSkip{0pt}

\acmJournal{TOG}
\acmYear{2020}\acmVolume{39}\acmNumber{4}\acmArticle{1}\acmMonth{7} \acmDOI{10.1145/3386569.3392391}

\setcopyright{acmcopyright}

\begin{document}

\newif\ifshowcomments
\showcommentsfalse


\ifshowcomments
\newcommand{\TODO}[1]{{\color{red}{[TODO: #1]}}}
\newcommand{\rz}[1]{{\color{magenta}#1}}
\newcommand{\hh}[1]{{\color[rgb]{0.1,0.46,0.9}#1}}
\newcommand{\rh}[1]{{\color[rgb]{0.5,0.0,0.5}#1}}
\newcommand{\ok}[1]{{\color[rgb]{0,0.5,0}#1}}
\newcommand{\change}[1]{{\color{blue}#1}}
\else
\newcommand{\TODO}[1]{#1}
\newcommand{\hh}[1]{#1}
\newcommand{\rh}[1]{#1}
\newcommand{\rz}[1]{#1}
\newcommand{\change}[1]{#1}
\newcommand{\ok}[1]{#1}
\fi


\newcommand{\gtop}{{\sc Graph2Plan}}

\title{Graph2Plan: Learning Floorplan Generation from Layout Graphs}

\author{Ruizhen Hu}
\affiliation{%
	\department{College of Computer Science \& Software Engineering}
	\institution{Shenzhen University}
}
\email{ruizhen.hu@gmail.com}

\author{Zeyu Huang}
\affiliation{%
	\institution{Shenzhen University}
}

\author{Yuhan Tang}
\affiliation{%
	\institution{Shenzhen University}
}

\author{Oliver van Kaick}
\affiliation{%
	\institution{Carleton University}
}

\author{Hao Zhang}
\affiliation{%
	\institution{Simon Fraser University}
}
\author{Hui Huang}
\authornote{Corresponding author: Hui Huang (hhzhiyan@gmail.com)}
\affiliation{%
	\department{College of Computer Science \& Software Engineering}
	\institution{Shenzhen University}
}

\renewcommand\shortauthors{R. Hu, Z. Huang, Y. Tang, O. van Kaick, H. Zhang, and H. Huang}

\begin{teaserfigure}
	\includegraphics[width=0.98\textwidth]{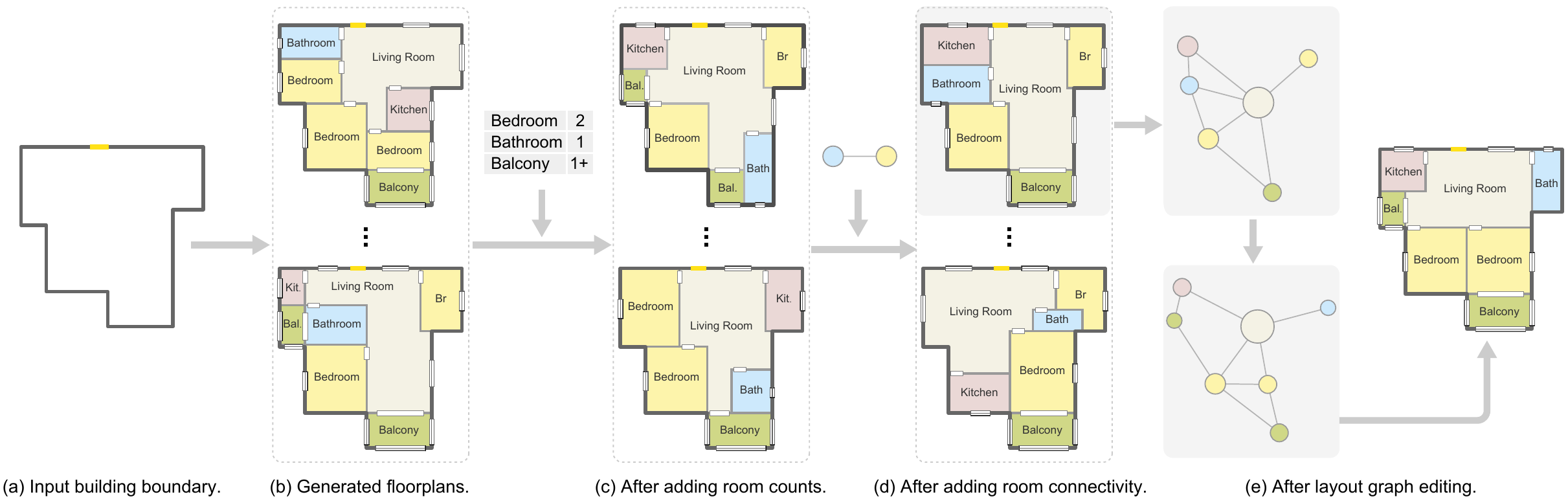}
	\centering
	\caption{Our deep neural network \gtop{} is a learning framework for automated floorplan generation from layout graphs. The trained network can generate floorplans based on an input building boundary only (a-b), like in previous works. In addition, we allow users to add a variety of constraints such as room counts (c), room connectivity (d), and other layout graph edits. Multiple generated floorplans which fulfill the input constraints are shown.}
	\label{fig:teaser}
\end{teaserfigure}

\input{abstract}

%
%


%
%


\maketitle

\input{intro}
\input{related}

\input{overview}

\input{method}
\input{results}

\input{conclusion}
\input{ack}

\bibliographystyle{ACM-Reference-Format}
\bibliography{references}

\end{document}

%% file: abstract.tex
\begin{abstract}
We introduce a learning framework for automated floorplan generation which combines generative modeling using deep 
neural networks and user-in-the-loop designs to enable human users to provide sparse design constraints. Such
constraints are represented by a {\em layout graph\/}. The core component of our learning framework is a deep neural 
network, \gtop{}, which converts a layout graph, along with a building boundary, into a floorplan that fulfills both the 
layout and boundary constraints.
Given an input building boundary, we allow a user to specify room counts and other layout constraints, which are used
to retrieve a set of floorplans, with their associated layout graphs, from a database.
For each retrieved layout graph, along with the input boundary, \gtop{} first generates a corresponding 
raster floorplan image, and then a refined set of boxes representing the rooms. \gtop{} is trained on RPLAN, a 
large-scale dataset consisting of 80K annotated floorplans. The network is
mainly based on convolutional processing over both the layout graph, via a graph neural network (GNN), and the input building 
boundary, as well as the raster floorplan images, via conventional image convolution. 
We demonstrate the quality and versatility of our floorplan generation framework in terms of its ability to cater to different
user inputs. We conduct both qualitative and quantitative evaluations, ablation studies, and comparisons with state-of-the-art approaches.
\end{abstract}

%% file: intro.tex
\section{Introduction}
\label{sec:intro}

One of the hottest recent trends in the field of designs, in particular, architectural designs, is the
adoption of AI and machine learning techniques. The volume and efficiency afforded by automated 
and AI-enabled generative models are expected to complement and enrich the architects' workflow, 
providing a profound and long-lasting impact on the design process. As one of the most 
fundamental elements of architecture, floor and building plans have drawn recent interests from
the computer graphics and vision community. Recent works on data-driven floorplan modeling
include raster-to-vector floorplan transformation~\cite{liu17}, floorplan reconstruction from 3D 
scans~\cite{liu18}, and floorplan or layout generation~\cite{wang19,wu19,ArchiGAN19}. 

\input{figures/overview}

In this paper, we introduce a learning framework for automated floorplan generation which combines 
generative modeling using deep neural networks and user-in-the-loop designs to enable human users
to provide initial sparse design constraints. We observe that user preferences for a target floorplan
often go beyond just a building boundary~\cite{wu19}; they may include more specific information such as room counts
or connectivity, e.g., the master bedroom should be next to the second (kid's) bedroom. Such constraints 
are best represented by a {\em layout graph\/}, much like scene graphs for image composition~\cite{johnson18}. 
Hence, the core component of our learning framework is a deep neural network, \gtop{}, which converts a 
layout graph, along with a building boundary, into a floorplan that fulfills both the layout and 
boundary constraints. Note that since the layout constraints are typically sparse, even
optional, they do not completely specify all rooms in the final floorplan. Consequently, the constraints
may lead to {\em one or more\/} suitable layout graphs, and in turn, multiple floorplans may be generated by our
method for the user to select and explore.


Given an input building/floor boundary, we first allow a user to optionally specify room counts and layout constraints, 
which are converted into a {\em query\/} layout graph to retrieve a set of floorplans from a dataset. If no such 
constraints are provided, then the building boundary serves as the query. In our work,
we utilize the large-scale floorplan dataset RPLAN~\cite{wu19} consisting of more than 80,000 human-designed
samples. The user can further refine the search by adjusting the query layout graph. 
\change{The advantages of this {\em retrieve-and-adjust\/} process are three-fold: as inputs to our floorplan generation 
network, the layout graphs associated with the retrieved floorplans carry both the user intent and design principles 
transferred from the RPLAN training set. At the same time, these graphs offer a concrete and intuive interface to facilitate user 
adjustment and refinement of the layout constraints.} 

Our deep generative neural network, \gtop{}, is also trained on the RPLAN dataset. For each retrieved layout graph, along 
with the input boundary, our network generates a corresponding raster floorplan {\em image\/} along with one bounding 
box for each generated room; see Figure~\ref{fig:overview} for an overview of our framework.
Note that while one box per room may not allow us to capture all possible room shapes, it is sufficient to cover the majority
as evidenced by the fact that over 93\% of the rooms in RPLAN can be represented as the intersection between their respective 
bounding boxes and the building boundary. The raster floorplan image is not the final output; it serves as a {\em global prior\/} 
and an intermediate representation to guide the individual box generation and refinement.

Figure~\ref{fig:network} shows the network architecture of \gtop, which is mainly based on convolutional processing over
both the layout graph, via a graph neural network (GNN), and the input building boundary, as well as the 
raster floorplan images, via conventional (image) convolutional neural networks (CNNs). Since the CNNs do not account
for {\em relational\/} information between the rooms, the resulting boxes from \gtop{} may not be well aligned. 
Hence, in the final step, we apply an optimization to align the room boxes and produce the final output as a {\em vectorized\/} 
floorplan.

We demonstrate with a \change{variety\/} of results, \change{as well as qualitative and quantitative evaluations\/}, that our learning-based 
framework is able to generate high-quality vectorized floorplans. \change{The intended users of our generative tool include floorplan 
designers for games, virtual reality, and large-scale planning projects, as well as end users, where they wish to {\em explore\/} 
early design intents, feasibility analyses, and mock-ups. The exploratory nature of these tasks would require flexibility and versatility
of the tool.} \change{Indeed, our framework provides varying degrees of control over the generative process. On one hand, since user input is optional, we can automatically generate a large variety of floorplans by retrieving template layout graphs only according to a room boundary and then executing \gtop{}. Thus, our framework can be used for mass floorplan generation, suitable for the creation of virtual worlds. On the other hand,} we also show that our tool enables users to guide the generation with design constraints, where the floorplans adequately adapt to the input boundary and the constraints, \change{leading to detailed floorplan mock-ups}. 
\change{We conduct a user study to show the plausibility of the generated floorplans. }
In addition, we perform an ablation study and comparison to a state-of-the-art approach to further evaluate our framework and the generated floorplans.

%% file: figures/overview.tex
\begin{figure*}[!t]
    \centering
    \includegraphics[width=\textwidth]{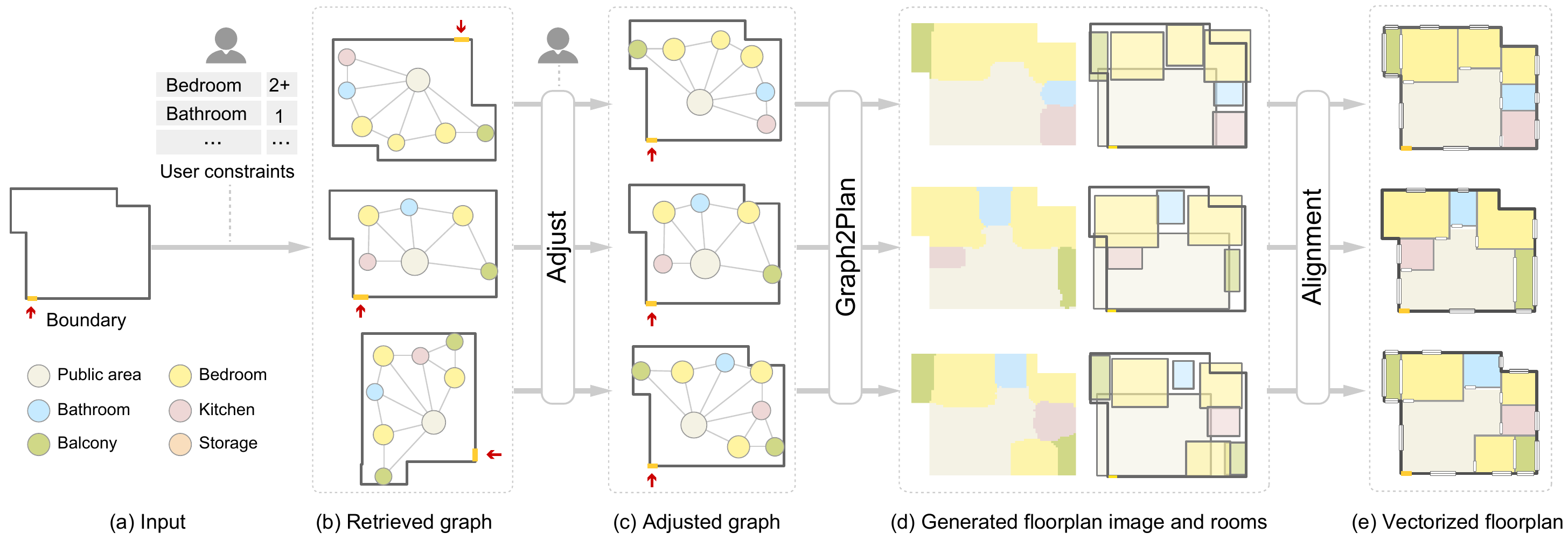}
\caption{
Overview of our framework for automated floorplan generation, which combines generative modeling using deep neural networks and user-in-the-loop design. 
(a) The user inputs a building boundary and can also specify constraints on room numbers, locations, and adjacencies. (b) We retrieve possible graph layouts from a dataset based on the user input. 
(c) Once graphs have been retrieved, we automatically transfer and adjust the graphs to the input boundary. The user can then interactively edit the room locations and adjacencies on the adjusted graphs. (d) The graphs guide our network in generating corresponding floorplans, encoded as a set of room bounding boxes and a floorplan raster image. (e) We post-process this output to obtain the final, vectorized floorplan.}
\label{fig:overview}
\end{figure*}

%% file: related.tex
\section{Related work}
\label{sec:related}


\noindent
In general, our work belongs to the topic of generative modeling of {\em structured arrangements\/}. Computer graphics research has introduced methods for the generation of a variety of structured arrangements and layouts, such as document layouts and clipart~\cite{li19}, urban layouts including street networks~\cite{yang13}, and game levels~\cite{hendrikx13}. As follows, we discuss the structured arrangements more closely related to our work, such as indoor scene synthesis, floorplan generation, and image composition.

\vspace{-3pt}

\paragraph{Indoor scene synthesis.}
The synthesis of indoor scenes typically involves the placement of furniture models from an existing database into a given room. As one of the first solutions to this problem, the method of Xu et al.~\shortcite{xu02} uses pseudo-physics and pre-defined placement constraints to arrange objects in a scene. Merrell et al.~ \shortcite{merrell11} introduce an interactive system that suggests furniture arrangements based on user constraints and interior design guidelines. More recent methods make use of data-driven approaches that learn from existing arrangements. For example, Fisher et al.~ \shortcite{fisher12} use a Bayesian network to synthesize new scenes from a given example, while subsequent work generates scenes from action maps predicted on an input scan, where the predictor is learned from training data~\cite{fisher15}. Zhao et al.~\shortcite{zhao16} synthesize scenes based on how objects interact with each other in an example scene. 
In contrast to floorplan generation, scene synthesis does not involve partitioning the input space into rooms and creation of room boundaries, but mainly the placement of objects into a room.

\vspace{-3pt}

\paragraph{Floorplan generation.}
Methods for floorplan generation typically take as input the outline of a
building and a set of user constraints, such as room sizes and
adjacencies between rooms, and propose a room layout that satisfies the
constraints, providing room locations and boundaries (walls). Earlier
efforts to address this problem used procedural or optimization methods
and manually-defined constraints. For example, Arvin and House~\shortcite{arvin02} generate indoor layouts with spring-systems whose equilibrium provides layouts close to the design objectives. Merrell et al.~\shortcite{merrell10} generate residential building layouts from high-level constraints with stochastic optimization and a learned Bayesian network. Rosser et al.~\shortcite{rosser17} follow up on this work by also considering the specification of a building outline and room characteristics. Rodrigues et al.~ \shortcite{rodrigues13part1,rodrigues13part2} generate building layouts from constraints with an evolutionary method.
Recently, Wu et al.~\shortcite{wu18} make use of mixed integer quadratic programming to generate building interior layouts.

A few approaches have also been proposed to generate other types of layouts related to floorplans. The method of Bao et al.~\shortcite{bao13} allows a user to explore exterior building layouts, while Feng et al.~\shortcite{feng16} optimize the layout of mid-scale spaces, such as malls and train stations, according to a crowd simulation. Ma et al.~\shortcite{ma14} generate game levels from an input graph with a divide-and-conquer strategy. In contrast to these approaches based on manual constraints or simple learning strategies such as Bayesian networks, our approach based on deep learning is able to implicitly learn the constraints for floorplan generation with the use of training data, leading to a more efficient method for generating floorplans.


\vspace{-3pt}

\paragraph{Deep learning for layout generation.}
Recent methods for layout generation make use of deep learning.
Wu et al.~\shortcite{wu19} introduce a deep network for the generation of floorplans of residential buildings. Given a building outline as input, the network predicts the locations of rooms and walls, which are then converted into a vector graphics format. The network is trained on a large-scale densely annotated dataset of floorplans of real residential buildings. One limitation of this approach is that the user has little control on the generated layout, besides the specification of the outline. Moreover, Chaillou~\shortcite{ArchiGAN19} proposes a three-step stack of deep networks for floorplan generation called ArchiGAN. The stack enables the generation of a building footprint, room floorplan, and furniture placement, all in the form of RGB images. The user is able to modify the input to each step by editing the images. However, no high-level control of the generation, such as room dimensions and specifications, is possible. In our work, we enable finer control of the generation by allowing the user to specify the desired layout at a high-level. Specifically, the user inputs a graph that provides the desired adjacencies between rooms and high-level properties of rooms. Also, besides the user intent, the layout graph carries design principles available from the training set, since the layout graphs are (at least partially) retrieved from the training set. Hence, our network incorporates a richer set of constraints and user goals when generating a floorplan.

Recently, deep neural networks have been developed by Li et al.~\shortcite{li2019_GRAINS} and Zhang et al.~\shortcite{zhang2019hybrid} for indoor scene generation, where the generative models can map a normal distribution to a distribution of object arrangements. Specifically, in GRAINS, Li et al.~\shortcite{li2019_GRAINS} train a variational autoencoder, based on recursive neural networks, to learn recursive object grouping rules in indoor environments. A new scene hierarchy can be generated by applying the trained decoder to a randomly drawn code. 

More closely related to our work, Wang et al.~\shortcite{wang19} introduce PlanIT, a \change{neurally-guided framework} for indoor scene generation according to input \change{scene} graphs. \change{Constrained by user-specified room specifications (e.g., walls and floors), as well as spatial and functional relations between furniture, their work first generates a scene graph and then instantiates the graph into a rectangular room via iterative furniture object insertion. In contrast, our work focuses on generating floorplans, which constitute partitioninigs of rectilinearly-shaped building boundaries into different types of rooms. Technically, their framework consists of a graph convolutional network (GCN) for scene graph 
generation, followed by an image-based and neurally-guided search procedure for scene instantiation. In our work, the layout graph is the result
of a retrieval and our \gtop{}~network is trained end-to-end to produce a floorplan image and the associated room boxes.}


Note that both GRAINS and PlanIT assume that the room boundaries are rectangular. In particular, alignment constraints are explicitly embedded into the GRAINS codes, with the four walls of a room serving as spatial references in all scene hierarchies to \change{define the alignments}. In contrast, our network \gtop{} works with arbitrary (rectilinear) building boundaries, \change{which are more general than rooms with four walls. With an indeterminate number of walls and many shape variations of the room boundaries, it is difficult to encode alignment constraints into the generative network and rely on simple heuristics to enforce them. This motivates our choice of using an additional refinement step for room alignment.} 



\vspace{-3pt}

\paragraph{Image composition from scene graphs.}
In computer vision, an important problem is to synthesize images by creating a scene composed of multiple objects. A recent solution to this problem is to derive the scene from a layout graph describing the location and names of objects that should appear in the image. Johnson et al.~\shortcite{johnson18} use graph convolution and an adversarial network for image generation from a graph. Ashual and Wolf~\shortcite{ashual19} improve the quality of the generated images by separating the input layout from the appearance of the target objects. We draw inspiration from these works by also guiding the floorplan generation with a graph. However, the problem setting is quite different, given that image generation involves the blending of objects rather than space partitioning.



\vspace{-3pt}

\paragraph{Graph generative models.}
Our method also relates to a line of work on representing graphs within deep networks. Recent approaches provide generative models of graphs, based on recurrent architectures~\cite{you18} or variational autoencoders~\cite{simonovsky18,grover18}. In our work, since our goal is not to generate graphs, but use the graphs to drive the generation of floorplans, we use the earlier graph neural network (GNN) model of Scarselli et al.~\shortcite{scarselli09} to encode graphs in the first layer of our network. The GNN enables to map a variety of graph types into a feature vector, which can then be further used by a deep network.

%% file: overview.tex
\section{Overview}
\label{sec:overview}

An overview of our floorplan generation framework is given in Figure~\ref{fig:overview}, with each component outlined below. Figure~\ref{fig:network} details
the architecture of the core component, the \gtop{} network, which is trained on the RPLAN dataset~\cite{wu19}.

\paragraph{Initial user input.}
The user first enters a building boundary for which the floorplan should be generated, and a set of constraints that guide the layout retrieval and floorplan generation (Figure~\ref{fig:overview}(a)). The user is allowed to provide as initial constraints the number of rooms of each type that should appear in the floorplan, the locations of specific rooms, and desired adjacencies between rooms. The rooms have types such as \emph{MasterRoom}, \emph{SecondRoom}, \emph{Kitchen}, etc. For example, a constraint could require 1 \emph{MasterRoom} and 1 \emph{BathRoom}.

\paragraph{Layout graphs.}
To obtain the layout graphs from user-specified constraints, we adopt a more traditional retrieve-and-adjust paradigm utilizing the large-scale floorplan dataset RPLAN~\cite{wu19}. \change{The key advantage of this approach is that the layout graphs are derived from real floorplans  incorporating human design principles.}
Specifically, in pre-processing, we first extract all the layout graphs from the dataset. Next, to retrieve relevant layout graphs, 
we filter the graphs based on the user constraints and then rank the graphs based on how well the source floorplan of the graph matches the boundary input by the user; \change{see Figure~\ref{fig:overview}(b)}. Note that, if the user did not provide any layout constraints, then the retrieval is based on the building boundary alone. After this step, the user can select one or more of the presented layout graphs as input for the next step. Our system also provides an interface that allows the user to interactively modify the retrieved graphs to further fulfill his/her constraints. The output of this step is a set of candidate layout graphs that guide the floorplan generation; \change{see Figure~\ref{fig:overview}(c)}. 

\paragraph{Floorplan generation.}
\change{Given a layout graph associated with a retrieved floorplan, our goal is to generate a new floorplan which instantiates the layout graph 
{\em within\/} the input building boundary. While the retrieved floorplan does provide a spatial realization of the layout graph, its boundary
is different from the input boundary. Retrofitting the layout graph into the new boundary corresponds to a ``{\em structural retargetting\/}'' problem. This problem is highly non-trivial without a clearly defined objective function, and simple heuristics to determine the room locations, sizes, and shapes will likely yield various conflicts. To this end, we train a deep neural network, coined \gtop, to solve the problem, where the network learns how to perform the retargeting based on design principles embedded in the floorplan dataset used as training data.}

The network takes as input a layout graph and building boundary, and generates a bounding box for each room; see Figure~\ref{fig:overview}(d). These bounding boxes may not be perfectly aligned and there may be some overlap between the boxes of different rooms. Thus, to ensure that the bounding boxes of all rooms can be composed into a valid floorplan, \gtop{} also predicts a raster floorplan image that assigns a single room label to each pixel inside the boundary. Finally, we solve an optimization problem that combines the raster floorplan image and bounding boxes to provide the final, valid alignment of room bounding boxes and a vectorized floorplan; see Figure~\ref{fig:overview}(e). \change{As discussed in Section~\ref{sec:related}, we require this additional refinement step since it is difficult to encode alignment constraints and enforce them during the floorplan generation with the input building boundaries taking on arbitrary shapes.}

%% file: method.tex
\section{Layout graph recommendation}

In this section, we explain how the layout graphs are extracted from the large-scale floorplan dataset RPLAN~\cite{wu19}, and how relevant graphs are retrieved based on user constraints. The retrieved layout graphs are first adjusted automatically to ensure that all the nodes are inside the user-provided building boundary. Then, the user can further edit the retrieved layout graphs, especially when some of the user constraints are not fully satisfied.

\subsection{Layout graph extraction}
\label{sec:extract}
The RPLAN dataset provides floorplans represented as raster images with semantic annotations at the pixel level~\cite{wu19}. Each image has four channels, which indicate the pixels that are inside/outside, on the boundary, and the room labels and instance indexes. To extract a layout graph for a given floorplan, we represent each room as a graph node and add an edge between any two nodes if the corresponding rooms are adjacent in the floorplan.

For each room node, we encode three pieces of information: room type, room location, and room size, illustrated in Figure~\ref{fig:info}(a). 
The room type can be one of the $N_t = 13$ following types: \emph{LivingRoom}, \emph{MasterRoom}, \emph{SecondRoom}, \emph{GuestRoom}, \emph{ChildRoom}, \emph{StudyRoom}, \emph{DiningRoom}, \emph{Bathroom}, \emph{Kitchen}, \emph{Balcony}, \emph{Storage}, \emph{Wall-in}, and \emph{Entrance}. 
For encoding the room location, we first divide the bounding box of the building boundary into a $K \times K$ grid, and then find the grid cell where the room center is located. We set $K=5$.
For encoding the room size, we compute the ratio between the room area and the whole building area. 

To find the adjacent room pairs, we first find all the interior doors encoded in the floorplan and then consider any rooms on the two sides of a door as adjacent pairs. 
Next, to find additional adjacent pairs, we check whether the distance between any two rooms is smaller than a given threshold relative to the room bounding box.
For each pair of adjacent rooms, we encode one piece of information: the pair's spatial relation type according to how the rooms are connected, which can be one of \emph{left of}, \emph{right of}, \emph{above}, \emph{below}, \emph{left-above}, \emph{right-above}, \emph{left-below}, \emph{right-below}, \emph{inside}, or \emph{outside}, illustrated in Figure~\ref{fig:info}(b). Since these spatial relations are directed, for each pair of adjacent rooms, we randomly sample one direction and assign the corresponding relation type to the edge.  


After extracting more than 80K layout graphs out of the 120K floorplans in the RPLAN dataset~\cite{wu19}, we find that the extracted subset shows enough variety of floorplan designs and can be used as an informative template set to help users in designing their own floorplan. 

\input{figures/info}

\subsection{Layout graph retrieval}

Users can specify layout constraints by providing the number of different room types that should appear in the floorplan, location of rooms, and adjacencies between rooms.
For simplicity, we cluster different kinds of bedrooms appearing in the dataset into one category, and only allow the users to add location and adjacency constraints on five types of rooms, as shown in the legend of Figure~\ref{fig:overview}(a). However, users have the option to specify room numbers for finer categories.
We filter out graphs that do not satisfy all of the user constraints, and then show the graphs to the user ranked by how well the boundary of the graphs' source floorplans matches the boundary provided by the user. If the user did not provide any constraints, we skip the filtering step.

We opt to rank graphs based on matching boundaries since buildings with similar boundaries will be more likely to have compatible floorplan designs that can be transferred to each other. However, even for two buildings with exactly the same boundary, different front door locations can lead to a significant change of the floorplan, implying that we need to take the front door location into consideration when comparing two boundaries. To achieve this, we first convert the 2D polygonal shape of each boundary into a 1D \emph{turning function}~\cite{Arkin91}, which records the accumulation of angles of each turning point (corner) of a polygon. Specifically, we start the sequence on one side of the front door and record the angles in a clockwise order; see Figure~\ref{fig:tf} for an example. Then, to compare two boundaries, we measure the distance between turning functions~\cite{Arkin91}. Figure~\ref{fig:adjust} shows one retrieval example.

\subsection{Layout graph adjustment}

We show each retrieved graph inside the input building boundary, so that the user can better analyze the fit of the layout to the boundary. To transfer a given layout graph to the input boundary, we first rotate the source plan of the graph so that its boundary aligns to the input boundary, rotating the graph as a consequence. Next, we transfer the nodes from the rotated graph to the boundary.

For the boundary alignment, since the boundary distance measure is based on the turning function starting from the front door, the front door serves as a reference point for the alignment. Thus, 
we first align the front doors of the two boundaries, which also prevents the front door from being blocked by any room.
To ensure that the transformed floorplan is still a valid floorplan with axis-aligned boundary edges, we constrain the transformation to be a rotation with $k \times 90$ degrees, where $k \in \mathbb{N}$. The variable $k$ is optimized so that the angle between the two front door directions is less than 45 degrees, where the front door direction is the vector connecting the center of the building bounding box to the center of the door. See Figure~\ref{fig:adjust}(a)-(c) for an example.

\input{figures/tf}

\input{figures/adjust}

Once the floorplan is aligned to the building boundary, we transfer the graph nodes. The position of each room node is encoded relative to the bounding box of the building boundary. Thus, we position a node inside the input boundary in the same relative position as it appears in the source floorplan. However, since the boundary of different buildings may differ significantly, this direct transfer of node positions may cause some of the nodes to fall outside of the building. Thus, the node locations need to be adjusted during their transfer. Since we represent a node position relative to a $5 \times 5$ grid, the adjustment can be restricted to the grid. For each room node falling outside of the building, we move it to the closest empty grid cell. If there is already a node inside this cell, we move the already existing node along the same direction to an adjacent cell. If needed, this process is iterated until the movement of nodes reaches the other side of the boundary. If the movement reached the last cell along the direction and there is already a node in the cell, we keep two nodes in the last cell. Note that this does not cause a problem since we still place nodes in their relative positions, and thus two nodes will not occupy the same position in space. See Figure~\ref{fig:adjust}(d) for an example of an adjusted graph.  Figure~\ref{fig:adjust}(e) shows the corresponding floorplan generated for the graph.


We present the input boundary and aligned graph to the user in an interactive interface, where the user can edit the retrieved graph and adapt it as needed. The user can add or delete room nodes and/or adjacency edges, or move nodes around to change the layout. 

\input{figures/network}

\section{Graph-based floorplan generation}

In this section, we introduce our \gtop~network for graph-based floorplan generation, and explain how the results are refined to provide the final vector representation of the generated floorplans.

\subsection{Network input and output}
\label{sec:data}
The input to the \gtop~network is the input building boundary $B$  and user-constrained layout graph $G=\{N, E\}$. The input boundary $B$ is represented as a $128\times128$ image with three binary channels encoding three masks, as in the work of Wu et al.~\shortcite{wu19}.
These masks capture the pixels that are inside the boundary, on the boundary, and on the entrance doors. 

For the layout graph, the nodes and edges are encoded in a similar manner as Ashual and Wolf~\shortcite{ashual19}. Specifically, each room $i$ in the floorplan is associated with a single node $n_i = [r_i, l_i, s_i]$, where $r_i \in \mathbb{R}^{d_1}$ is a learned encoding of the room category, $l_i \in \{0,1\}^{d_2}$ is a location vector, and $s_i \in \{0,1\}^{d_3}$ is a size vector. The room category embedding $r_i$ is one of $c$ possible embedding vectors, $c=13$ being the number of room categories, and $r_i$ is set according to the category of room $i$. This embedding is learned as part of the network training,
and the embedding size $d_1$ is set arbitrarily to 128. $d_2$ is set to be 25 to denote a coarse image location using a $5 \times 5$ grid, and $d_3$ is set to be 10 to denote the size of a room using different scales. The edge information $e_{ij} \in \mathbb{R}^{d_1}$ also encodes a learned embedding for the relations between the nodes. In other words, the values of $e_{ij}$ are taken from a learned dictionary with ten possible values, each associated with one type of pairwise relation.

The output of the network is a $128\times128$ floorplan image $\mathcal{I}$ and two sets of room bounding boxes: an initial set of boxes $\{\mathcal{B}_i^0\}$, and a refined set $\{\mathcal{B}_i^1\}$, with  $\mathcal{B}_i = [x_i,y_i,w_i,h_i]$ being the predicted bounding box for room $n_i$.


\input{figures/box_refine}

\subsection{Network architecture}
A diagram of the full architecture of our \gtop~network is shown in Figure~\ref{fig:network}. The layout graph $G$ is first passed to a graph neural network (GNN)~\cite{scarselli09} that embeds each room in the layout into a feature space. The boundary features are obtained through a conventional encoder applied to $B$, whose output is concatenated with each of the room features. Then, for each room, the concatenated features are used to generate a corresponding bounding box through the network denoted as \emph{Box}. All the predicted room boxes are then used to guide the composition of room features for the generation of the floorplan image $\mathcal{I}$ through a cascaded refinement network (CRN)~\cite{chen17}. 
\change{For overlapping regions, we sum up the corresponding room features.}
To make use of the global information gathered in the floorplan image $I$, inspired by works on object detection~\cite{girshick15}, we consider the previously predicted boxes as regions of interest (RoIs) and refine each box one by one through an additional network, which we denote \emph{BoxRefineNet}.

The network architecture of \emph{BoxRefineNet} is shown in Figure~\ref{fig:box_refine}. More specifically, \emph{BoxRefineNet} first processes the whole image with several convolutional and max pooling layers to produce a feature map. Then, for each room box, a RoI pooling layer extracts a fixed-length feature vector from the feature map and initial bounding box. This feature vector is then concatenated with the room features and fed into a new \emph{Box} network consisting of a sequence of fully connected layers that output a refined box position and size.

\subsection{Loss functions}
To train the \gtop~network, we design suitable loss functions to account for each type of output. Let us recall that the network predicts a raster floorplan image $\mathcal{I}$, initial boxes $\{\mathcal{B}_i^0\}$, and refined boxes $\{\mathcal{B}_i^1\}$. The loss function is then defined as:
\begin{equation}
L =	L_{\text{pix}}(\mathcal{I}) + L_{\text{reg}}(\{\mathcal{B}_i^0\}) + L_{\text{geo}}(\{\mathcal{B}_i^0\}) +  L_{\text{reg}}(\{\mathcal{B}_i^1\}),
\label{eq:loss}
\end{equation}	
where $L_{\text{pix}}(\mathcal{I})$ is the image loss that is simply defined as the cross entropy of the generated floorplan image and the ground truth floorplan image, $L_{\text{reg}}(\{\mathcal{B}_i\})$ is the regression loss, which penalizes the $L_1$ difference between ground-truth and predicted boxes, and $L_{\text{geo}}(\{\mathcal{B}_i^0\})$ is the geometric loss, which ensures the geometric consistency among boxes and between boxes and the input boundary. Note that the geometric loss is only applied to the initial boxes; since the refined boxes are only adjusted locally, optimizing their geometric consistency leads to little improvement.

The geometric loss of the initial boxes $\{\mathcal{B}_i^0\}$ is defined as:
\begin{equation}
\begin{split}
L_{\text{geo}}(\{\mathcal{B}_i^0\}) =&\ L_{\text{coverage}}(\{\mathcal{B}_i^0\})
+ L_{\text{interior}}(\{\mathcal{B}_i^0\}) \\
& + L_{\text{mutex}}(\{\mathcal{B}_i^0\})
+ L_{\text{match}}(\{\mathcal{B}_i^0\}),
\end{split}
\label{eq:loss_geo}
\end{equation}	
where $L_{\text{coverage}}$ and $L_{\text{interior}}$ both constrain the spatial consistency between the boundary $B$ and the room bounding box set $\{\mathcal{B}_i\}$, $L_{\text{mutex}}$ constrains the spatial consistency between any two room boxes $\mathcal{B}_i$ and $\mathcal{B}_j$, and $L_{\text{match}}(\{\mathcal{B}_i\})$ ensures that the predicted boxes match the ground-truth boxes. The first three terms are inspired by the work of Sun et al.~\shortcite{sun19} and ensure the proper coverage of the building interior by the boxes. We extend their formulation with the last term that compares the boxes to the ground-truth, and thus ensures that the prediction of box locations and dimensions is also improved during training. Figure~\ref{fig:loss} illustrates the geometric loss.

Before giving more details about the terms of the geometric loss, we first define two distance functions, $d_{\text{in}}(p, \mathcal{B})$ to measure the coverage of a point $p$ by a box $\mathcal{B}$, and $d_{\text{out}}(p, \mathcal{B})$  to measure how far a point $p$ is from a box $\mathcal{B}$:
\[ d_{\text{in}}(p, \mathcal{B})=
\begin{cases}
0,  & \text{if}~p \in \Omega_{\text{in}}(\mathcal{B});\\
\min_{q \in \Omega_{\text{bd}}(\mathcal{B})} ||p-q||, & \text{otherwise}.
\end{cases} \]
\[ d_{\text{out}}(p, \mathcal{B})=
\begin{cases}
0,  & \text{if}~p \notin \Omega_{\text{in}}(\mathcal{B});\\
\min_{q \in \Omega_{\text{bd}}(\mathcal{B})} ||p-q||, & \text{otherwise}.
\end{cases} \]
The sets $\Omega_{\text{in}}(\mathcal{B})$ and $\Omega_{\text{bd}}(\mathcal{B}) $  denote the interior and boundary of $\mathcal{B}$, respectively. We define the terms of the geometric loss as follows.

\input{figures/loss}

\paragraph{Coverage loss.} The input building $B$ should be fully covered by the union of all the room boxes. Specifically, any point $p \in \Omega_{\text{in}}(B)$  should be covered by at least one room box. Thus, the coverage loss is defined as follows:
\begin{equation}
L_{\text{coverage}}(\{\mathcal{B}_i\}) = \frac{\sum_{p \in \Omega_{\text{in}}(B)}{\min_{i}d_{\text{in}}(p, \mathcal{B}_i)^2}}{|\Omega_{\text{in}}(B)|},
\end{equation}	
where $|\Omega_{\text{in}}|$ is the number of pixels in the set $\Omega_{\text{in}}(B)$.

\paragraph{Interior loss.}  Each of the room bounding boxes should be located inside of the boundary bounding box $\hat{B}$. Thus, the interior loss can be defined as follows: 
\begin{equation}
L_{\text{interior}}(\{\mathcal{B}_i\}) = \frac{\sum_{i} \sum_{p \in \Omega_{\text{in}}(\mathcal{B}_i) }{d_{\text{in}}(p, \hat{B})^2}}{\sum_{i} |\Omega_{\text{in}}(\mathcal{B}_i)| }.
\end{equation}	

\paragraph{Mutex loss.} The overlap between room boxes should be as small as possible so that the rooms are compactly distributed inside the building. Thus, the mutex loss can be defined as follows:
\begin{equation}
L_{\text{mutex}}(\{\mathcal{B}_i\}) = \frac{\sum_{i} \sum_{p \in \Omega_{\text{in}}(\mathcal{B}_i)} \sum_{j \neq i}{d_{\text{out}}(p,  \mathcal{B}_j)^2}}{\sum_{i} \sum_{p \in \Omega_{\text{in}}(\mathcal{B}_i) } \sum_{j \neq i}1}.
\end{equation}


\paragraph{Match loss.}  Each of the room bounding boxes $\mathcal{B}_i$ should cover the same region as the corresponding ground-truth box $\mathcal{B}_i^*$, that is, $\mathcal{B}_i$ should be located inside of $\mathcal{B}_i^*$ and $\mathcal{B}_i^*$ should be located inside of $\mathcal{B}_i$. Thus, the match loss can be defined as follows: 
\begin{equation}
\begin{split}
L_{\text{match}}(\{\mathcal{B}_i\}) =&\ \frac{\sum_{i} \sum_{p \in \Omega_{\text{in}}(\mathcal{B}_i) }{d_{\text{in}}(p, \mathcal{B}_i^*)^2}}{\sum_{i} |\Omega_{\text{in}}(\mathcal{B}_i)| } + \\
& \frac{\sum_{i} \sum_{p \in \Omega_{\text{in}}(\mathcal{B}_i^*) }{d_{\text{in}}(p, \mathcal{B}_i)^2}}{\sum_{i} |\Omega_{\text{in}}(\mathcal{B}_i^*)| }.
\end{split}
\end{equation}	

\subsection{Room alignment and floorplan vectorization}
\label{sec:align}
The final output of \gtop{} is a raster floorplan image and one bounding box for each room. An issue that may occur with the output boxes is that they may not be well aligned and some boxes may overlap in certain regions. Thus, in the final vectorization step, we use the raster floorplan image to determine the room label assignment in the regions with overlap, i.e., we determine the layer ordering of different rooms and at the same time improve the alignment of room boundaries with a heuristic method.  


We first align the rooms with the building boundary and then align adjacent rooms with each other. More specifically, for each edge of a room box, we find the nearest boundary edge with the same orientation, i.e., horizontal or vertical, and align the box edge with the boundary edge if their distance is smaller than a given threshold $\tau$. 
Furthermore, we align adjacent room pairs based on their encoded spatial relation in the layout graph. For example, if room A is on the left of room B, we snap the right edge of room A with the left edge of room B. In addition, we also snap the top and bottom edges of the two rooms if they are also close enough according to the threshold $\tau$, since it is better for rooms that are placed side by side in a floorplan to have aligned walls in order to minimize the number of corners. 
Note that one room box may be adjacent to different room boxes. Thus, the edges have to be updated several times. To avoid breaking previously refined alignments, we set a flag to indicate whether a box edge has already been updated or not. If any of the edges has already been updated, it remains fixed and we align the other edge to the fixed edge.
If both edges are not fixed, we update them to their average position. Figure ~\ref{fig:align} shows an example of room alignment.

\input{figures/align}

\input{figures/order}

Moreover, we need to determine the room category label for regions covered by overlapping boxes \change{and determine the drawing order of those room boxes}. To achieve that, for each pair of rooms, we check if they overlap and use the generated floorplan image $\mathcal{I}$ to determine their relative order. In more detail, for each room pair, we count the number of pixels inside the region with overlap that has the label of each room, and determine that the room with the smaller count should be drawn first. 
\change{If two rooms get the same vote, the one with larger area will be drawn first.}
Following this procedure, we build a graph by adding one node to represent each room, and one directed edge from room $R_2$ to room $R_1$, if $R_1$ and $R_2$ overlap and $R_1$ should be drawn before $R_2$. Then, our goal is to find an order of all the graph nodes satisfying the ordering constraints imposed by the directed edges. To find such an ordering, \change{we first randomly select any node with \emph{outdegree} equal to 0, and then remove all edges pointing to the node from the rest of the graph. Here, the \emph{outdegree} is the number of directed edges starting from the node which equals the number of rooms that need to be drawn before the room node.} We continue removing nodes with outdegree equal to 0 until the graph becomes empty.
Note that, if there is a loop in the graph, we cannot find a linear order for the nodes in the loop. Thus, we randomly select the node with minimal outdegree to delete and break the loop. \change{Figure ~\ref{fig:order} shows an example of determining room ordering.}

Finally, we add windows and internal doors to the floorplan based on the
heuristics proposed by Wu et al.~\shortcite{wu19}, i.e., add doors
between connected rooms and windows along exterior wall segments.


%% file: figures/info.tex
\begin{figure}[!t]
    \centering
    \includegraphics[width=0.48\textwidth]{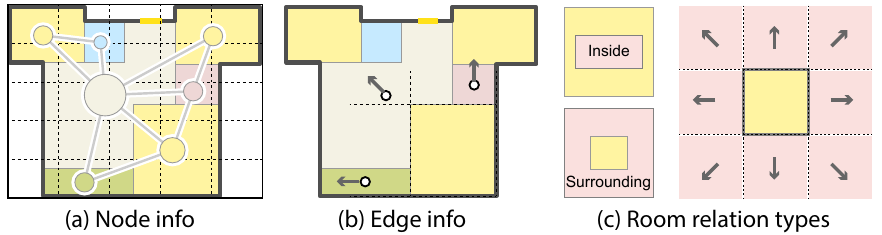}
\caption{Information encoded in the layout graph: (a) The nodes encode the room type, size, and location relative to a $5 \times 5$ grid. (b) The edges encode the spatial relation type between two rooms. In the example, the spatial relations are relative to the large yellow room on the bottom-right of the floorplan. (c) All the room relation types that we consider in our framework.}
\label{fig:info}
\end{figure}

%% file: figures/tf.tex
\begin{figure}[!t]
    \centering
    \includegraphics[width=0.48\textwidth]{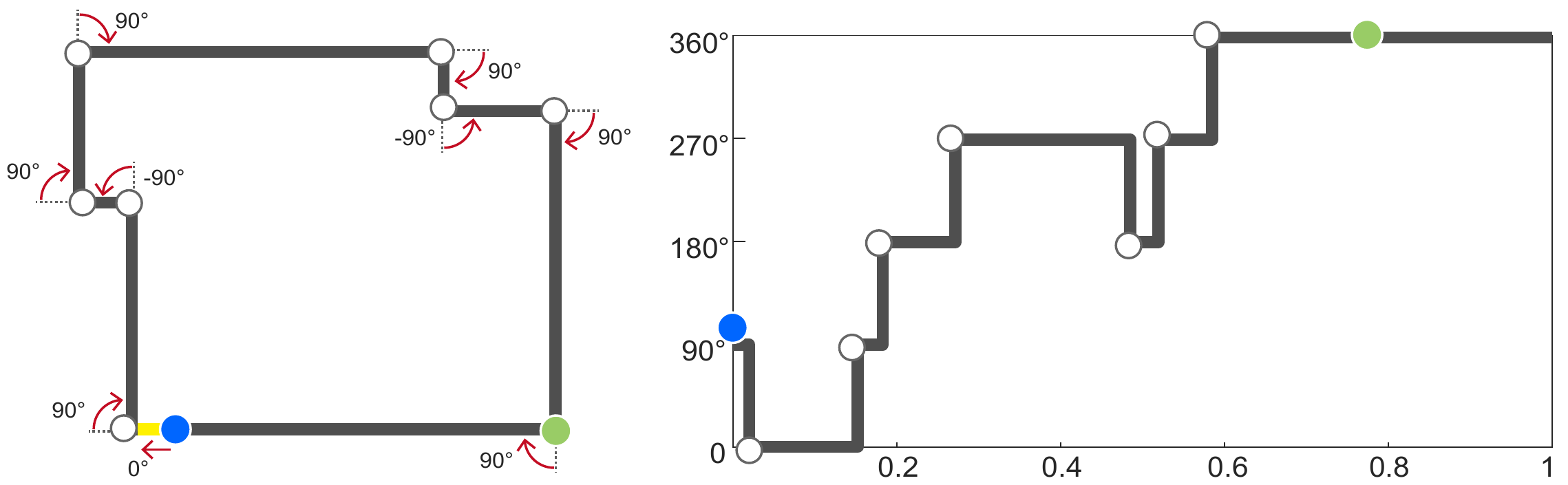}
\caption{Turning function of a building boundary: the encoding starts from the blue point besides the door highlighted in yellow, and proceeds clockwise around the polygon, registering the angle sum at each corner, \change{where each edge is normalized by the sum of all edges}.}
\label{fig:tf}
\end{figure}

%% file: figures/adjust.tex
\begin{figure}[!t]
    \centering
    \includegraphics[width=0.48\textwidth]{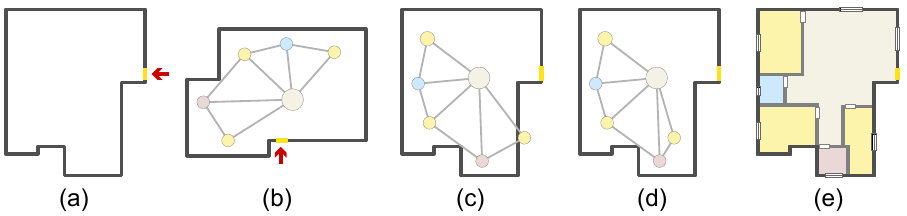}
\caption{Example of retrieval and adjustment of a layout graph: (a) Input boundary, (b) Retrieved graph, (c) Transferred graph, (d) Adjusted graph, (e)~Generated floorplan. }
\label{fig:adjust}
\end{figure}

%% file: figures/network.tex
\begin{figure*}[!t]
    \centering
    \includegraphics[width=\textwidth]{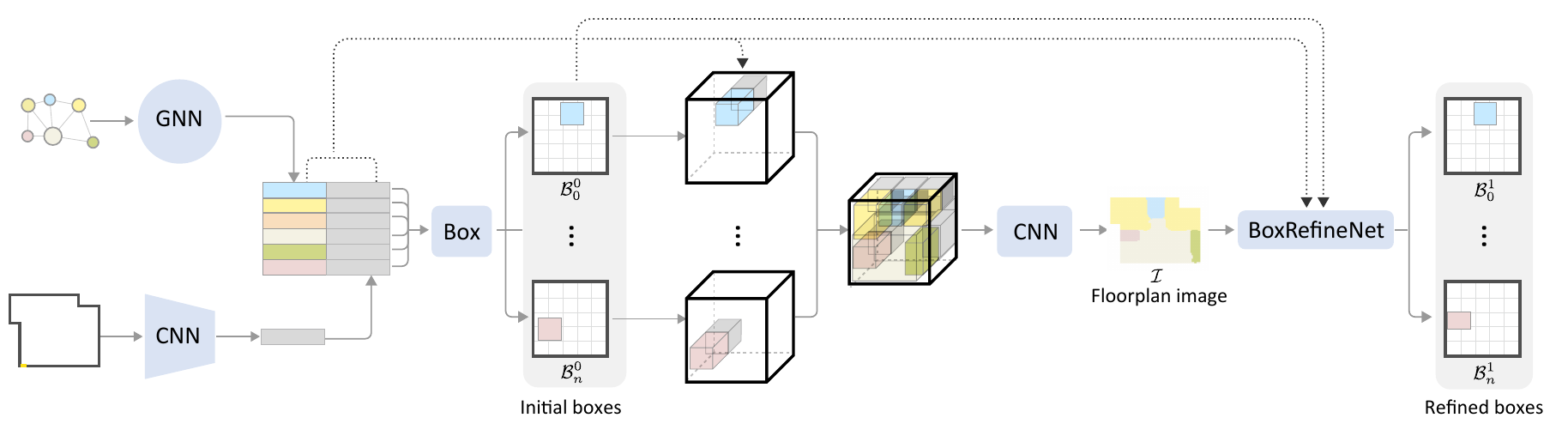}
\caption{
Architecture of our \gtop~network. The network takes as input a layout graph and building boundary, and outputs initial room bounding boxes $\{\mathcal{B}_i^0\}$, refined room boxes $\{\mathcal{B}_i^1\}$, and a raster image $\mathcal{I}$ of the floorplan. The processing is carried out with a graph neural network (GNN), convolutional neural network (CNN), fully-connected layers (Box), and a BoxRefineNet (detailed in Figure~\ref{fig:box_refine}).}
\label{fig:network}
\end{figure*}

%% file: figures/box_refine.tex
\begin{figure}[!t]
    \centering
    \includegraphics[width=0.48\textwidth]{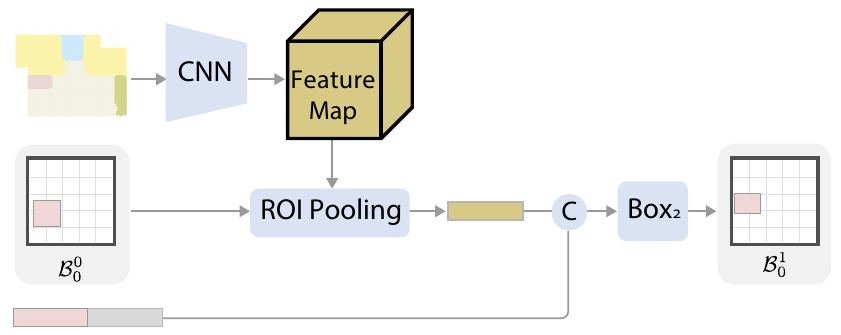}
\caption{
Network architecture of the BoxRefineNet: the network
produces a feature map of the input boundary with a CNN. A RoI Pooling layer then extracts a feature vector from the feature map and input bounding box. The feature vector is concatenated with the room features, and fed into fully-connected layers \change{(\emph{$\text{Box}_2$})} that output a refined box position and size.} 
\label{fig:box_refine}
\end{figure}

%% file: figures/loss.tex
\begin{figure}[!t]
    \centering
    \includegraphics[width=0.48\textwidth]{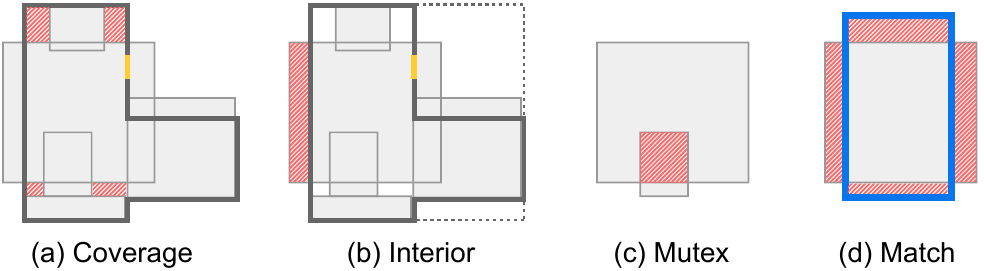}
\caption{
Illustration of the geometric loss: (a) Coverage term that ensures that the building is covered by the union of all boxes, (b) Interior term that constrains boxes to the interior of the boundary, (c) Mutex term that prevents overlap among room boxes, and (d) Match term constraining boxes to cover the same region as their
ground-truth counterpart. The input boundary is shown in black color, predicted boxes are in grey, ground-truth boxes have a blue boundary, and the red areas are used to calculate the loss terms.}
\label{fig:loss}
\end{figure}

%% file: figures/align.tex
\begin{figure}[!t]
    \centering
    \includegraphics[width=0.48\textwidth]{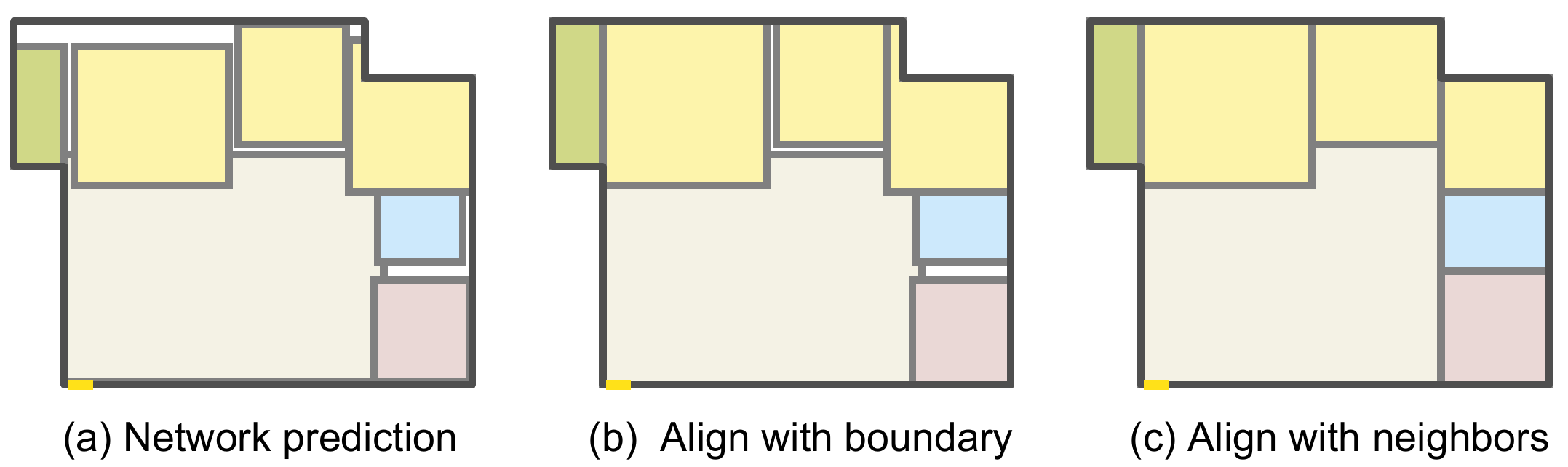}
\caption{
Example of room alignment. (a) Given the boxes predicted by the network, (b) we align the boxes to the building boundary, and then (c) align adjacent boxes to each other.
}
\label{fig:align}
\end{figure}

%% file: figures/order.tex
\begin{figure}[!t]
    \centering
    \includegraphics[width=0.48\textwidth]{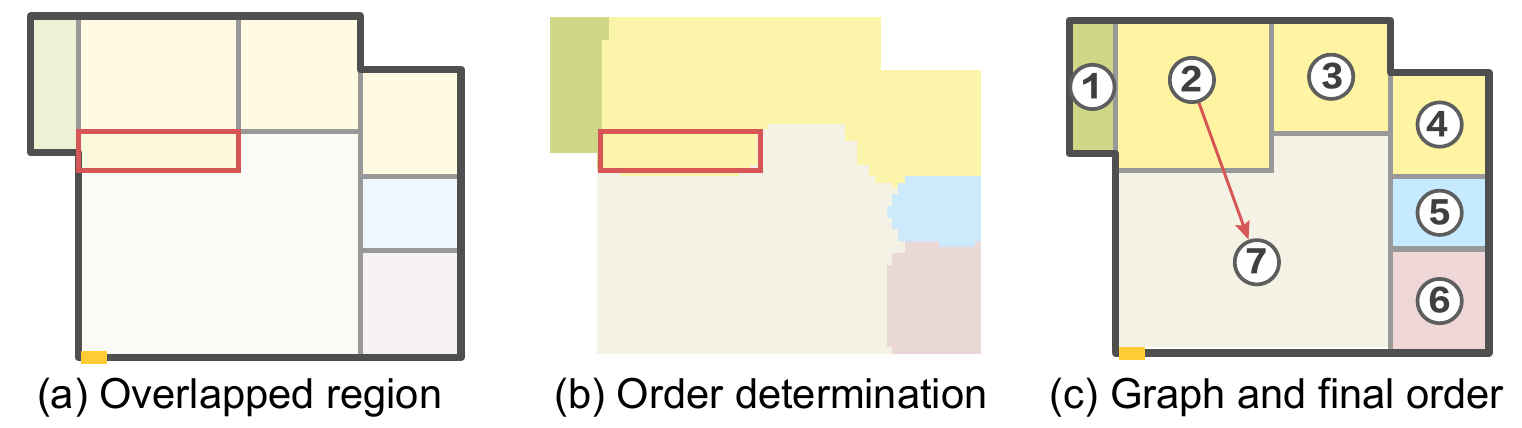}
\caption{
\change{Example of determining room ordering. (a) Given the room boxes with one overlapped region indicated by the red box, (b) we check the corresponding region in the generated floorplan image $\mathcal{I}$ to determine the relative drawing order of the overlapping room pair. Then, (c) a corresponding directed edge is added to the room graph to indicate that one room needs to be drawn before the other.  In this example, the only constraint is that room 7 should be drawn before room 1.   Then, we find the final draw ordering of the rooms respecting the constraints.}
}
\label{fig:order}
\end{figure}

%% file: results.tex
\section{Results and evaluation}

We discuss the creation of training data for \gtop, and then evaluate our results qualitatively and quantitatively.

\subsection{Training data preparation}
The training data for the \gtop~network is derived from RPLAN~\cite{wu19}. The input data is composed of the building boundary and layout graph of each floorplan, while the output data is composed of the room bounding boxes and raster image. The input boundary of each floorplan can be easily extracted from the annotated floorplan data and the layout graph extraction is explained in Section~\ref{sec:extract}. The room bounding boxes can also be easily extracted from the annotated floorplans. However, these boxes cannot be used directly since our method only generates rooms without walls, while the boxes extracted from RPLAN have gaps between rooms after the walls are removed. To fill the gaps between rooms, we use our alignment method described in Section~\ref{sec:align}.
\change{Note that, after alignment, 99\% of the areas of original room boxes are inside the refined room boxes, and the average IoU between the original room boxes and refined room boxes is 0.86. Thus, the refined room boxes used as our ground-truth are quite consistent with the original data.}
Once we refined the room box locations, we use them to regenerate the raster floorplan image $\mathcal{I}$ without interior walls. 
The training-validation-test split of the data is 70\%--15\%--15\%.


\subsection{Network training}
\label{sec:training}

Our network is trained progressively since the later stages of the network depend on good initial predictions of room boxes. Thus, we train the network in three steps. In the first step, we train the portion of the network that predicts initial room boxes from the input boundary and graph, using the loss terms $L_{\text{reg}}(\{\mathcal{B}_i^0\})$ and $L_{\text{geo}}(\{\mathcal{B}_i^0\})$ from Eq.~\ref{eq:loss}. Next, we also train the generation of the raster image, adding the corresponding loss. Finally, we also train the {\em BoxRefineNet}, using all the terms of Eq.~\ref{eq:loss}. 

\subsection{User interface}


\input{figures/interface}

Figure~\ref{fig:interface} shows a snapshot of our interface for floorplan design. Once an input boundary is loaded on the left panel, the user can specify room constraints with the dropdown boxes on the top bar or by creating a partial graph on the left panel. Then, the best matching floorplans are retrieved and shown in the middle panel. The user can inspect a retrieved result by clicking on it and seeing a zoomed-in version of the floorplan and layout graph on the right panel. If the user likes the suggested design, they can click the transfer button to transfer the graph to the input boundary and  automatically adjust the node locations so that they fit into the boundary. The user can further edit the transferred graph by moving nodes around, deleting or adding edges, and even deleting or adding new nodes. Once the user is satisfied with the layout graph, they can click a button to generate a corresponding floorplan. After that, the process can be iterated by further editing the graph and regenerating the floorplan.

\change{Given an input boundary, it takes 99ms to retrieve the list of similar floorplans from the database, 11ms for layout graph transfer, 68ms for floorplan generation using the pre-trained network, and 200ms for the post-processing step. Thus, in total, generating a floorplan from an input boundary takes less than 0.4 seconds. In contrast, it takes about 4 seconds to generate a floorplan with the method of Wu et al.~\shortcite{wu19}. Note that the initialization of our system to open the interface and load all the data takes around 10 seconds.}

%
%

\input{figures/comp_b_c}
\subsection{Qualitative results}
Our method accepts different types and numbers of user constraints and generates the corresponding floorplans. 
Figure~\ref{fig:comp_b_c} shows a set of example results generated from different boundary inputs and different user constraints. Each row shows the results of different layout constraints applied to the same boundary, while each column shows the results obtained when the same layout constraints are applied to different boundaries. The selected constraints are the desired numbers of three room types: bedroom, bathroom, and balcony. The corresponding constraint on the room number is shown on the bottom of each column. 


By examining each row, we see how the generated floorplans satisfy the given room number constraints and adapt to the input boundary. Different numbers of bedrooms, bathrooms, and balconies are generated based on the constraints, and the location of these rooms changes to best conform the floorplan to the input boundary. Note how the balconies in green usually have two or three faces on the building boundary, reflecting the typical balcony design in apartments. Thus, their location changes depending on the input boundary. All of the floorplans also have a living room, as it is present in buildings with complex boundaries like these, and can have additional rooms.

From the results shown on each column, we see how the same number of rooms with the same type distribute differently inside the buildings with different boundaries. For example, the two bathrooms in the third column are sometimes adjacent to each other and sometimes not, but always adjacent to bedrooms. In the fifth column, balconies are never adjacent to each other, and usually appear at different locations in the building, showing a range of variation in the resulting floorplans.

Moreover, for each set of layout constraints, our method can retrieve more than one floorplan that satisfies all the constraints, and then automatically transfer the graph to the input boundary to guide the floorplan generation. Figure~\ref{fig:multi} shows multiple results generated for the same boundary and same constraints, i.e., the floorplans should have at least 2 bedrooms, 1 bathroom, and 1 balcony. \change{Note the diversity of the results even when generated from the same input boundary and set of layout constraints.}

\change{Furthermore, different front door locations can lead to significantly different floorplans, even with a building boundary of the same shape. Figure~\ref{fig:door} shows example results for the same input boundary shape but with different front door locations. In this example, we did not add any user constraints so that the floorplan with the most similar boundary defined by the turning function starting from the door is retrieved and used to define the graph. Note how the room arrangement changes with the different door locations.}

\change{To test how our method deals with complex inputs, we show example results with complex input boundaries in the last two rows of both Figure~\ref{fig:comp_b_c} and \ref{fig:multi}, where the boundaries have many corners.  
Furthermore, we also show example results with complex input constraints in the second row of Figure~\ref{fig:door}, where we retrieve layout graphs with the maximal number of nodes (8) in the dataset to constrain the generation. 
We see that our method provides reasonable results for both complex input boundaries and input constraints.}

%

\input{figures/multi}

\input{figures/door}

\paragraph{Graph adjustment}
Figure~\ref{fig:edit} shows a few example results before and after the user edits the graph. From (a) to (b), the user broke the link between the pink and yellow nodes on the top-left corner and then moved the pink node to the bottom-left corner. Note how, after regenerating the floorplan, the left part of the room layout has been changed while the right part has been kept the same. From (b) to (c), the user added a new green node (for Balcony) between two rooms on the top, and a new room is generated in the corresponding space. From (c) to (d), the user added an edge between the new green node and the top-left yellow node to enforce the two rooms to be aligned. Note how the corresponding changes appear in the floorplan.

\input{figures/edit}

\input{figures/comp}
\paragraph{Comparison to Wu et al.~\shortcite{wu19}}
To compare with the floorplan generation method of Wu et al.~\shortcite{wu19}, we generate floorplans only from the input boundaries without taking any user constraints. In this way, the floorplan with the most similar boundary is retrieved and the corresponding layout graph is transferred for floorplan generation. Figure~\ref{fig:comp} shows the comparison. We see that our floorplans are comparable in the amount of detail and quality to those obtained by Wu et al.~\shortcite{wu19}. In addition, our method can generate a variety of floorplans from a single boundary, with different numbers, types, and arrangements of rooms. In general, the generated floorplans capture design principles that are embedded in the training data. With our method, the users can also edit the retrieved layout graphs to fine-tune the floorplans according to their design intent.


%
%

\input{figures/user_study}
\input{figures/ablation}

\change{
\subsection{Qualitative evalution}

We evaluate the quality of a sample of floorplans generated by \gtop{} in a user study where we asked users to compare generated floorplans with ground-truth (GT) floorplans taken from the test set. For the study, we randomly selected 20 GT floorplans and set the room types and numbers of the GT floorplans as the constraints for retrieval and floorplan generation with our method. 
In the study, we showed the generated floorplan besides the corresponding GT, and asked users which floorplan is more plausible without revealing each source. Users could select either of them as more plausible or ``Cannot tell''. Note that the boundary, room type and numbers are the same in the two floorplans, while the only difference is the room layout. We presented the floorplans to the users in random order. We also generated two additional filter tasks comparing GT floorplans to floorplans generated from randomly placed boxes, which are obviously less plausible than GT floorplans. Only the selections of users who passed the filter tasks were considered valid responses. 

We asked 30 participants to do the user study, all of which are graduate students in computer science, and collected 600 answers in total. 
The votes for the options ``GT/ours/cannot tell'' are 304/218/78, respectively. Thus, for 50.7\% of the questions, participants think that the ground-truth floorplans are more plausible than our results, while for 49.3\%, participants think that our results are at least as plausible, if not more plausible, than the GT, which indicates the high plausibility of our results.
Figure~\ref{fig:user_study} shows examples of user selections for different floorplans. In (a), users tend to like more regular rooms; in (b), users prefer larger rooms (kitchen and bedrooms); in (c), the only difference between the two layouts is the size of the kitchen, which is either aligned with one of the boundary corners or with an adjacent bathroom, where users could not decide which floorplan is more plausible. Moreover, as shown in (b) and (c), the room layout generated by our method is similar to the ground truth, which indicates that implicitly there exist design constraints in the dataset which are transferred from the retrieved floorplan to the generated floorplan.
}

\subsection{Quantitative evalution}
\paragraph{\gtop~evaluation}
To evaluate our \gtop~network quantitatively, we take each building boundary and corresponding layout graph from the test set, and compare the room bounding boxes generated by \gtop{} to the ground-truth boxes extracted from the source floorplan. We compare the boxes with the Intersection over Union (IoU) measure. On average, our method obtains IoU scores of 0.65, which indicate good prediction, given that the optimum value for IOU is 1. 

Figure~\ref{fig:ablation} shows examples where we visually compare the predicted boxes (Setting 5 column) to the ground-truth (GT columns). We see that, overall, the predicted room boxes are similar to the ground-truth, appearing in the same location as the ground-truth boxes and with similar size. However, small differences close to the room boundaries can occur, e.g., some rooms are slightly shorter than in the ground-truth.

\paragraph{Ablation study}
To justify our network design, we perform four ablation studies with different combinations of network components and loss functions, and compare them to the results with the full network and loss function, resulting in five evaluation settings. The settings and corresponding IoUs are: 
\begin{itemize}
    \item Setting 1: box regression only, using $L_{\text{reg}}(\{\mathcal{B}^0_i\})$. IoU = 0.43.
    \item Setting 2: box regression + geometry constraint, using $L_{\text{reg}}(\{\mathcal{B}^0_i\})$ + $L_{\text{geo}}(\{\mathcal{B}^0_i\})$. IoU = 0.47.
    \item Setting 3: box regression + image composition, using $L_{\text{reg}}(\{\mathcal{B}^0_i\})$ + $L_{\text{pix}}(\mathcal{I})$. IoU = 0.54. 
    \item Setting 4: box regression + geometry constraint + image composition, using $L_{\text{reg}}(\{\mathcal{B}^0_i\}) + L_{\text{geo}}(\{\mathcal{B}^0_i\})+	L_{\text{pix}}(\mathcal{I})$. IoU = 0.56.
    \item Setting 5: our complete network with box regression + geometry constraint + image composition + box refinement, using $L_{\text{reg}}(\{\mathcal{B}^0_i\}) + L_{\text{geo}}(\{\mathcal{B}^0_i\})+	L_{\text{pix}}(\mathcal{I}) + L_{\text{reg}}(\{\mathcal{B}_i^1\})$. IoU = 0.66.
\end{itemize}

Note that Setting 3 can be seen as the adaptation of the method of Ashual and Wolf~\shortcite{ashual19} to our problem. In the comparison between Setting 1 and 3, we see that using the predicted box to compose the raster image for global guidance improves the generated floorplans. Examples can be found in Figure~\ref{fig:ablation}. We see that a few rooms can go missing without the image guidance (rows 2 and 4). 

When comparing Settings 1 and 2, and 3 and 4, we see that adding the geometric loss on the initial boxes can slightly improve the box IoU. Note that the first three terms of the geometric loss, i.e., coverage, interior, and mutex term, do not take any ground-truth information into consideration, and just aim to make the box distribution inside the boundary look better, i.e., less overlap and more coverage. Thus, it is reasonable that these three terms alone do not improve the IoU performance much. However, with these three terms, the box distribution looks better in general, as shown in Figure~\ref{fig:ablation}. E.g., in Setting 3, more boxes are extending to the exterior of the buildings, which are adjusted in Setting 4.

Finally, when comparing our complete network (Setting 5) to all the other settings, we see that the performance is improved significantly with the box refinement step. We see in Figure~\ref{fig:ablation} that the room boxes are significantly tighter after the refinement. 

\change{
To further justify the necessity of the box refinement component in the network and show the effectiveness of our post-processing step for room alignment and floorplan vectorization, we also compare the generation results obtained before the post-processing step to those obtained after this step. 
We find that the IoU before the post-processing step is initially 0.54 and then 0.66 with refinement, while the IoU after the post-processing step is initially 0.75 and then 0.80 with refinement. 
From these results, we conclude that: 1) the vectorization step works well; 2) the refinement component in the Graph2Plan network is useful and the results before and after vectorization are both improved. 
}

%% file: figures/interface.tex
\begin{figure}[!t]
    \centering
    \includegraphics[width=0.48\textwidth]{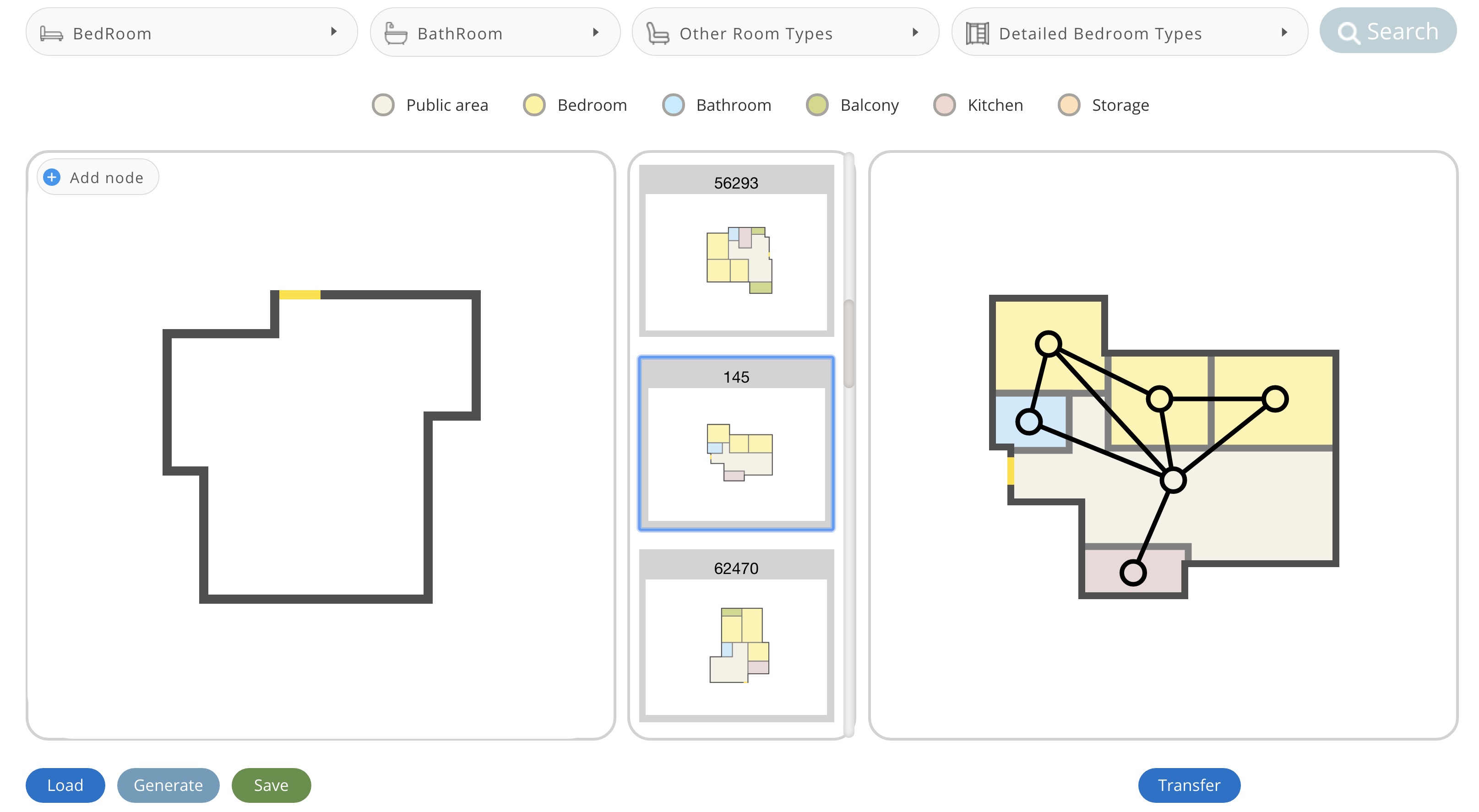}
\caption{
Our interface for user-in-the-loop design of floorplans. Left: input boundary. Middle: retrieved floorplans. Right: floorplan and layout graph of a retrieved result. Please refer to the text for more details.
}
\label{fig:interface}
\end{figure}

%% file: figures/comp_b_c.tex
\begin{figure}[!t]
    \centering
    \includegraphics[width=0.485\textwidth]{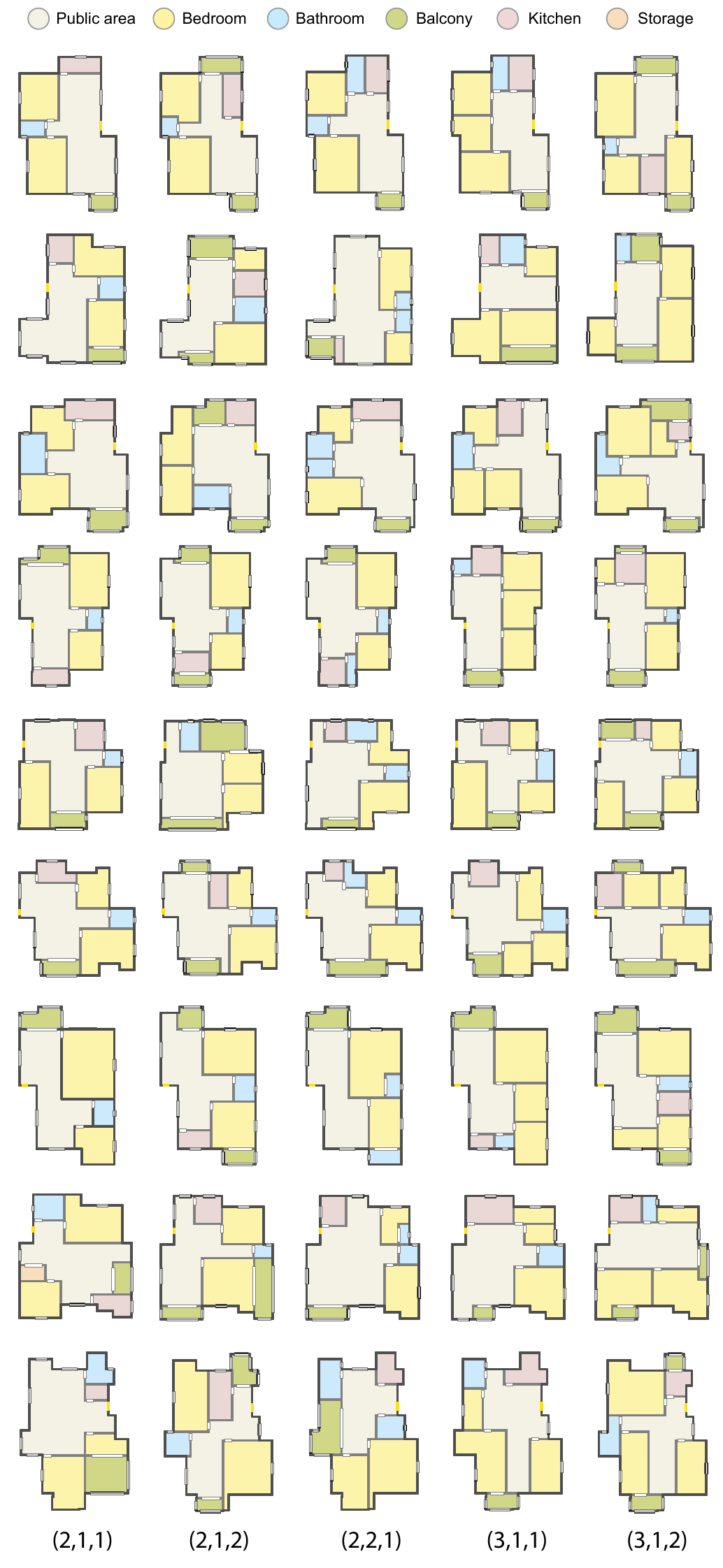}
\caption{Gallery of floorplans generated with our method. The rows show results generated for different input boundaries, while the columns show results generated for different constraints. The constraints are the desired number of three room types: bedroom (in yellow), bathroom (blue), and balcony (green). The constraints are shown on the bottom of each column.}
\label{fig:comp_b_c}
\end{figure}

%% file: figures/multi.tex
\begin{figure}[!t]
    \centering
    \includegraphics[width=0.48\textwidth]{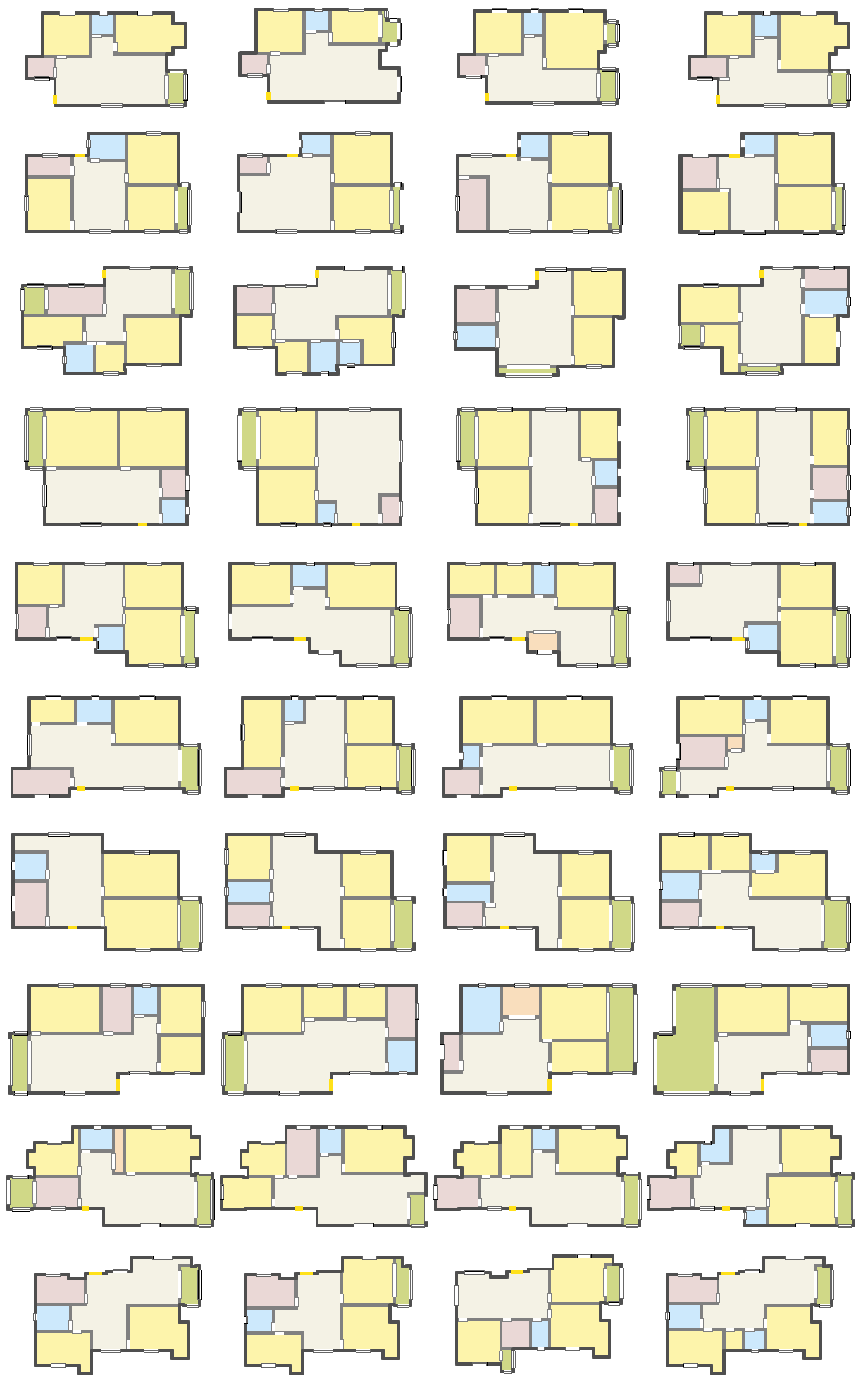}
\caption{Each row shows multiple floorplans generated from the same boundary and constraints (at least 2 bedrooms, 1 bathroom, and 1 balcony).}
\label{fig:multi}
\end{figure}

%% file: figures/door.tex
\begin{figure}[!t]
    \centering
    \includegraphics[width=0.48\textwidth]{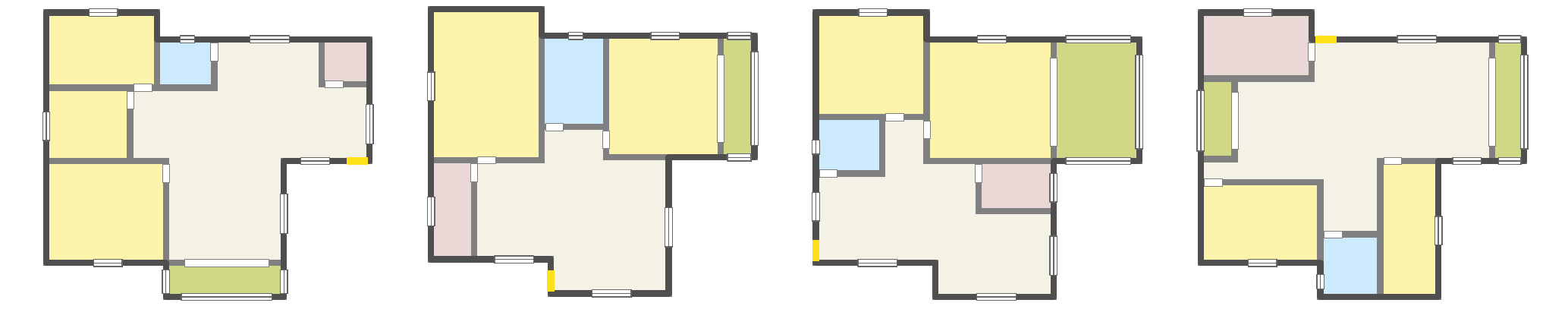}     
     \includegraphics[width=0.48\textwidth]{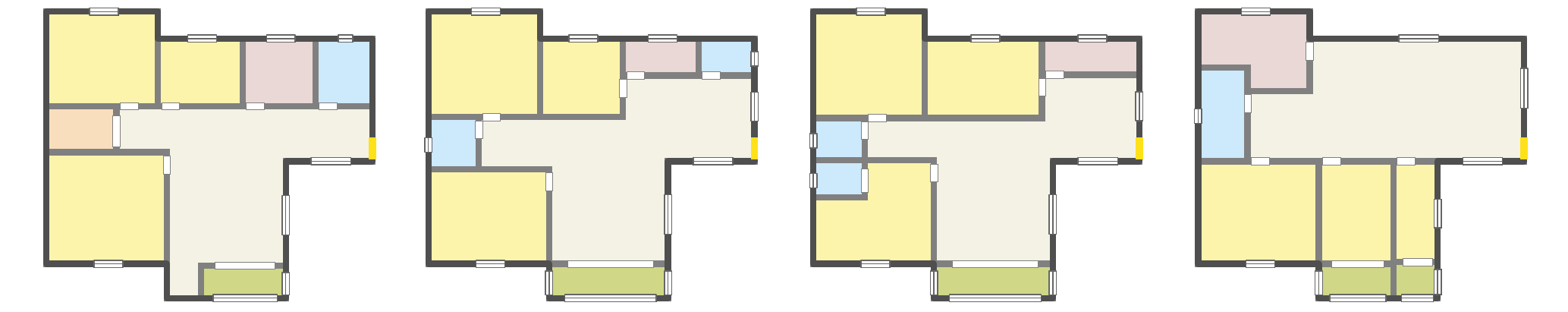}
\caption{\change{The first row shows floorplans generated for input boundaries with the same shape but different front door locations (in yellow). The second row shows floorplans generated for the same boundary but with more complex layout constraints by setting the required number of rooms to 8. }}
\label{fig:door}
\end{figure}

%% file: figures/edit.tex
\begin{figure}[!t]
    \centering
    \includegraphics[width=0.48\textwidth]{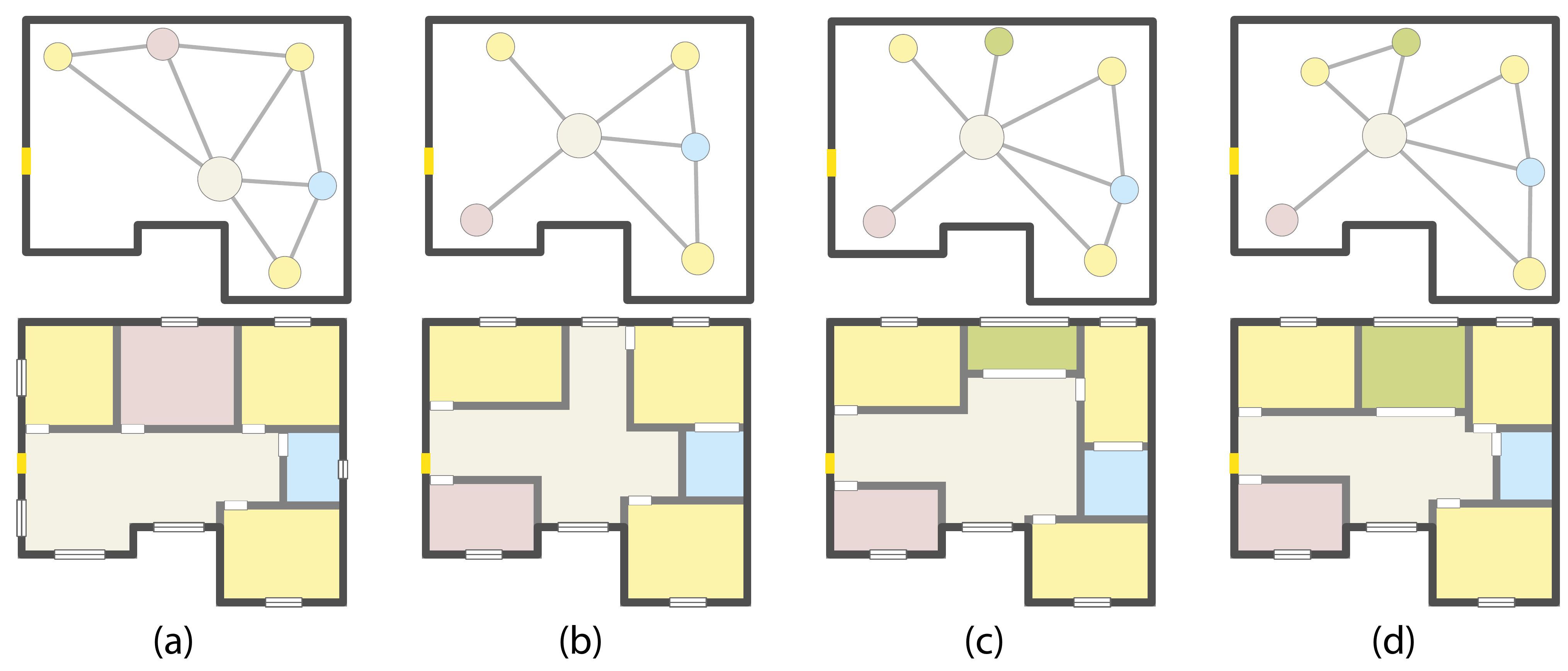}
\caption{Floorplan update after the user adjusted the layout graph. (a) Initial graph and generated floorplan. (b) Node moved and edge removed. (c) New node added. (d) New edge added.}
\label{fig:edit}
\end{figure}

%% file: figures/comp.tex
\begin{figure}[!t]
    \centering
    \includegraphics[width=0.49\textwidth]{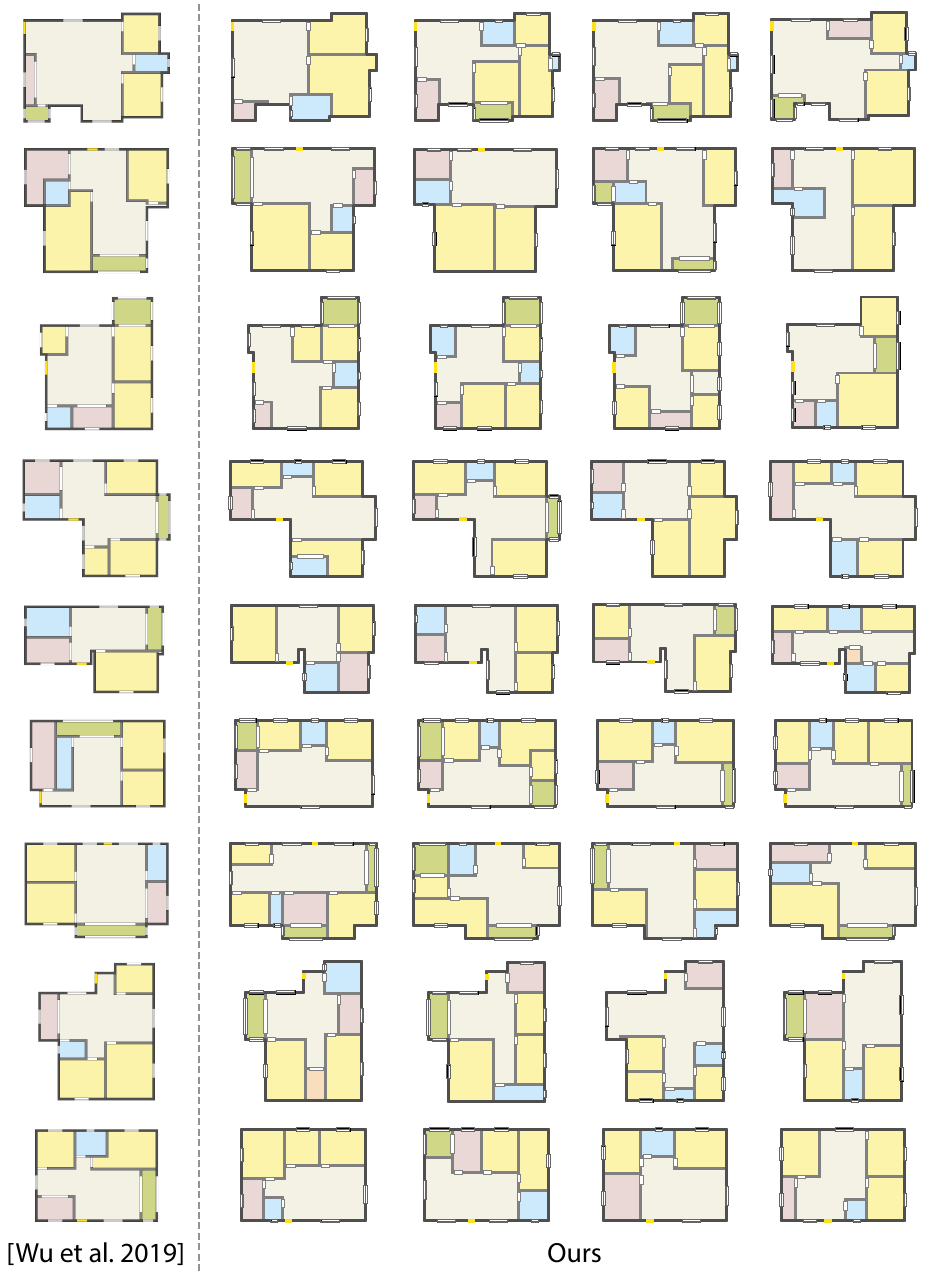}
\caption{Comparison to Wu et al.~\shortcite{wu19} by applying their and our method on the same input boundary. Note that we can generate multiple floorplans for a single input boundary, with a variety of room numbers and arrangements.}
\label{fig:comp}
\end{figure}

%% file: figures/user_study.tex
\begin{figure}[!t]
    \centering
    \includegraphics[width=0.48\textwidth]{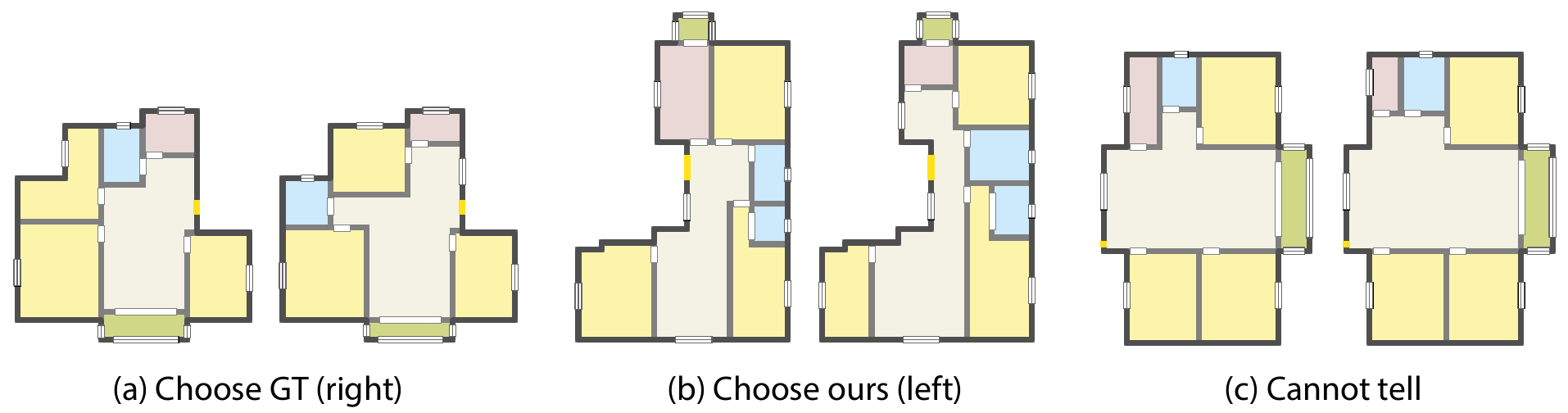}     
\caption{\change{Examples of floorplans presented in our user study. In each example, our result is shown on the left and the ground-truth (GT) floorplan is on the right, while the user selection is indicated in the caption.}}
\label{fig:user_study}
\end{figure}

%% file: figures/ablation.tex
\begin{figure*}[!t]
    \centering
    \includegraphics[width=0.98\textwidth]{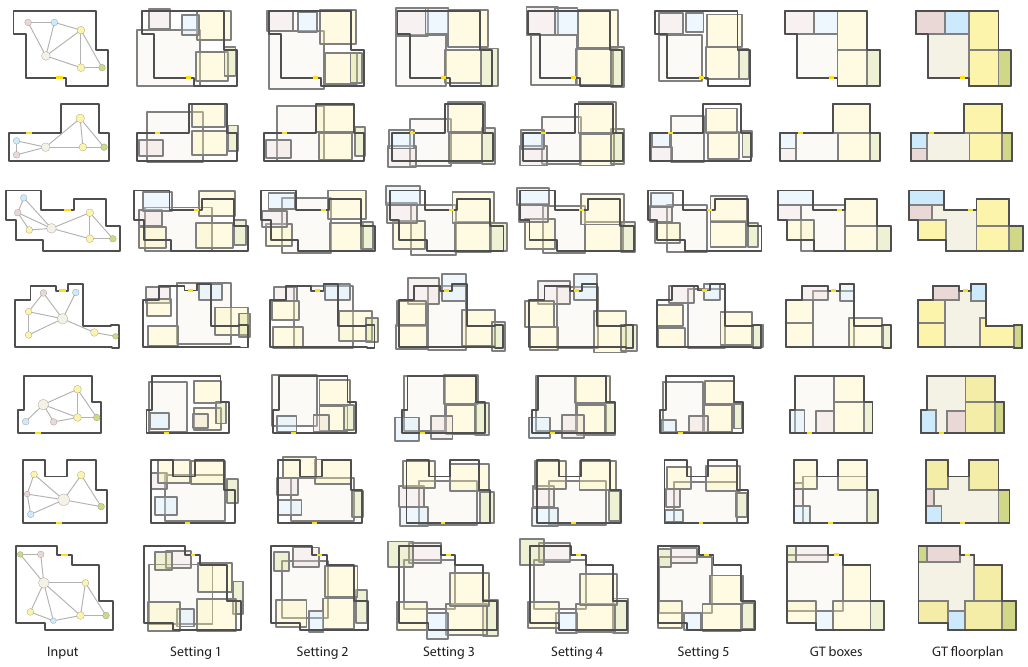}
\caption{Ablation study demonstrating the role of each component of the network in generating high-quality floorplans. Each setting evaluates the isolated effect of one or more terms of the loss function, which correspond to specific components of the network. Please refer to the text for details on each setting.}
\label{fig:ablation}
\end{figure*}

%% file: conclusion.tex
\section{Conclusion and future work}
 \label{sec:future}

We introduce the first deep learning framework for floorplan generation that enables user-in-the-loop modeling. The users can specify their design goals with constraints that guide the retrieval of layout graphs from a large dataset of floorplans, and can further refine the constraints by editing the layout graphs. The layout graphs specify the desired numbers and types of rooms along with the room adjacencies, and directly guide the floorplan generation. We demonstrated with a series of experiments that this framework allows users to generate a variety of floorplans from the same input boundary, and fine-tune the results by editing the layout graphs. In addition, a quantitative evaluation shows that the floorplans are similar to training examples and thus also tend to follow the design principles learned from the dataset of floorplans.

\input{figures/failure}

\paragraph{Limitations.}
As this work is a first step in the direction of user-guide floorplan
generation, it has certain limitations. Although the layout
graphs model the user preferences in terms of desired room types and their
location, the types of constraints encoded in the graphs are limited. For example, the graphs do not model accessibility criteria or functionality considerations of the floorplans. Thus, these considerations are also not captured by the learned network. In addition, the user cannot specify that certain rooms should not be adjacent to each other, or adjacent to certain features of the input boundary. Specifically, features such as interior doors and windows are not captured by the current model. Moreover, the alignment of predicted rooms and vectorization is dealt with in a post-processing step, and thus is not part of the core learning framework. \change{Finally, if the layout graph is retrieved from a small dataset where the obtained  boundary is significanly different from the input one, our method may fail to generate a floorplan that satisfies all the constraints; see Figure~\ref{fig:failure} for an example.}

%
%

\paragraph{Future work.}
Aside from addressing the current limitations, directions for future work include extending the scope and capabilities of our user-guided generation framework. For example, our framework could be adapted to guide the synthesis of furniture based on preferences captured with a part layout graph. 
In addition, the framework could be enhanced by allowing users to perform more complex manipulations with the layout graphs, such as retrieving other graphs similar to a query graph or combining graphs. 
\change{Moreover, it is possible to incorporate other types of user constraints into our system. For example, the positioning of support walls could be passed to the network as part of the input boundary. In that case, it would be possible to add another loss function for support walls and create the appropriate data to train the network.}
Finally, we could also learn or derive a structural model of the floorplans from the generated designs, to enable users to fine-tune the structure of the final room layouts, especially the wall locations.

%

%% file: figures/failure.tex
\begin{figure}[!t]
    \centering
    \includegraphics[width=0.48\textwidth]{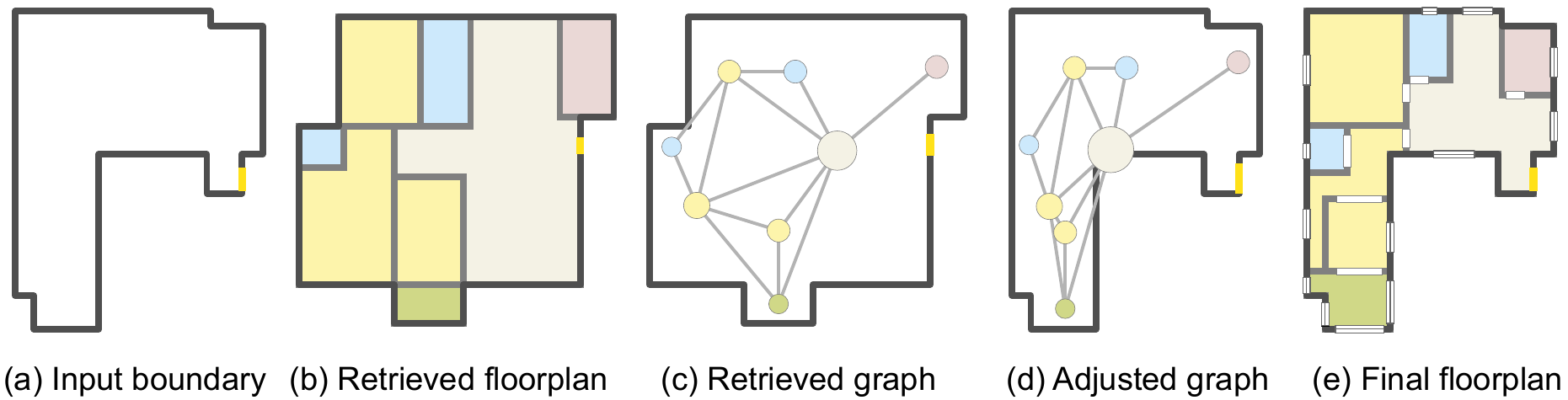}
\caption{\change{Failure case. Given the input boundary (a), if the retrieved floorplan (b) has quite a different boundary, then the corresponding retrieved graph (c) needs to be sufficiently adjusted (d) to fit into the input boundary to properly guide the floorplan generation. (e) As a result, in this example, room nodes are distributed closely together and the final generated rooms have overlaps. In addition, some of the relationship constraints given by the edges cannot be satisfied.}}
\label{fig:failure}
\end{figure}

%% file: ack.tex
\section*{Acknowledgements}
We thank the anonymous reviewers for their valuable comments. This work was supported in part by NSFC (61872250, 61861130365, 61761146002), GD Higher Education Key Program (2018KZDXM058), GD Science and Technology Program (2015A030312015), LHTD (20170003), NSERC Canada (611370, 611649, 2015-05407), gift funds from Adobe, National Engineering Laboratory for Big Data System Computing Technology, and Guangdong Laboratory of Artificial Intelligence and Digital Economy (SZ), Shenzhen University.

%% file: main.bbl

\begin{thebibliography}{34}


\ifx \showCODEN    \undefined \def \showCODEN     #1{\unskip}     \fi
\ifx \showDOI      \undefined \def \showDOI       #1{#1}\fi
\ifx \showISBNx    \undefined \def \showISBNx     #1{\unskip}     \fi
\ifx \showISBNxiii \undefined \def \showISBNxiii  #1{\unskip}     \fi
\ifx \showISSN     \undefined \def \showISSN      #1{\unskip}     \fi
\ifx \showLCCN     \undefined \def \showLCCN      #1{\unskip}     \fi
\ifx \shownote     \undefined \def \shownote      #1{#1}          \fi
\ifx \showarticletitle \undefined \def \showarticletitle #1{#1}   \fi
\ifx \showURL      \undefined \def \showURL       {\relax}        \fi
\providecommand\bibfield[2]{#2}
\providecommand\bibinfo[2]{#2}
\providecommand\natexlab[1]{#1}
\providecommand\showeprint[2][]{arXiv:#2}

\bibitem[\protect\citeauthoryear{{Arkin}, {Chew}, {Huttenlocher}, {Kedem}, and
  {Mitchell}}{{Arkin} et~al\mbox{.}}{1991}]%
        {Arkin91}
\bibfield{author}{\bibinfo{person}{E.~M. {Arkin}}, \bibinfo{person}{L.~P.
  {Chew}}, \bibinfo{person}{D.~P. {Huttenlocher}}, \bibinfo{person}{K.
  {Kedem}}, {and} \bibinfo{person}{J.~S.~B. {Mitchell}}.}
  \bibinfo{year}{1991}\natexlab{}.
\newblock \showarticletitle{An efficiently computable metric for comparing
  polygonal shapes}.
\newblock \bibinfo{journal}{\emph{IEEE Trans. Pattern Analysis \& Machine
  Intelligence}} \bibinfo{volume}{13}, \bibinfo{number}{3}
  (\bibinfo{date}{March} \bibinfo{year}{1991}), \bibinfo{pages}{209--216}.
\newblock
\showISSN{1939-3539}
\urldef\tempurl%
\url{https://doi.org/10.1109/34.75509}
\showDOI{\tempurl}


\bibitem[\protect\citeauthoryear{Arvin and House}{Arvin and House}{2002}]%
        {arvin02}
\bibfield{author}{\bibinfo{person}{Scott~A. Arvin} {and}
  \bibinfo{person}{Donald~H. House}.} \bibinfo{year}{2002}\natexlab{}.
\newblock \showarticletitle{Modeling architectural design objectives in
  physically based space planning}.
\newblock \bibinfo{journal}{\emph{Automation in Construction}}
  \bibinfo{volume}{11}, \bibinfo{number}{2} (\bibinfo{year}{2002}),
  \bibinfo{pages}{213--225}.
\newblock


\bibitem[\protect\citeauthoryear{Ashual and Wolf}{Ashual and Wolf}{2019}]%
        {ashual19}
\bibfield{author}{\bibinfo{person}{Oron Ashual} {and} \bibinfo{person}{Lior
  Wolf}.} \bibinfo{year}{2019}\natexlab{}.
\newblock \showarticletitle{Specifying Object Attributes and Relations in
  Interactive Scene Generation}. In \bibinfo{booktitle}{\emph{Proc. Int. Conf.
  on Computer Vision}}.
\newblock


\bibitem[\protect\citeauthoryear{Bao, Yan, Mitra, and Wonka}{Bao
  et~al\mbox{.}}{2013}]%
        {bao13}
\bibfield{author}{\bibinfo{person}{Fan Bao}, \bibinfo{person}{Dong-Ming Yan},
  \bibinfo{person}{Niloy~J. Mitra}, {and} \bibinfo{person}{Peter Wonka}.}
  \bibinfo{year}{2013}\natexlab{}.
\newblock \showarticletitle{Generating and Exploring Good Building Layouts}.
\newblock \bibinfo{journal}{\emph{ACM Trans. on Graphics (Proc. SIGGRAPH)}}
  \bibinfo{volume}{32}, \bibinfo{number}{4} (\bibinfo{year}{2013}),
  \bibinfo{pages}{122:1--122:10}.
\newblock


\bibitem[\protect\citeauthoryear{Chaillou}{Chaillou}{2019}]%
        {ArchiGAN19}
\bibfield{author}{\bibinfo{person}{Stanislas Chaillou}.}
  \bibinfo{year}{2019}\natexlab{}.
\newblock \emph{\bibinfo{title}{AI + Architecture: Towards a New Approach}}.
\newblock \bibinfo{thesistype}{Master's\ thesis}. \bibinfo{school}{Harvard
  School of Design}.
\newblock


\bibitem[\protect\citeauthoryear{Chen and Koltun}{Chen and Koltun}{2017}]%
        {chen17}
\bibfield{author}{\bibinfo{person}{Qifeng Chen} {and} \bibinfo{person}{Vladlen
  Koltun}.} \bibinfo{year}{2017}\natexlab{}.
\newblock \showarticletitle{Photographic image synthesis with cascaded
  refinement networks}. In \bibinfo{booktitle}{\emph{Proc. Int. Conf. on
  Computer Vision}}. \bibinfo{pages}{1511--1520}.
\newblock


\bibitem[\protect\citeauthoryear{Feng, Yu, Yeung, Yin, and Zhou}{Feng
  et~al\mbox{.}}{2016}]%
        {feng16}
\bibfield{author}{\bibinfo{person}{Tian Feng}, \bibinfo{person}{Lap-Fai Yu},
  \bibinfo{person}{Sai-Kit Yeung}, \bibinfo{person}{KangKang Yin}, {and}
  \bibinfo{person}{Kun Zhou}.} \bibinfo{year}{2016}\natexlab{}.
\newblock \showarticletitle{Crowd-driven Mid-scale Layout Design}.
\newblock \bibinfo{journal}{\emph{ACM Trans. on Graphics (Proc. SIGGRAPH)}}
  \bibinfo{volume}{35}, \bibinfo{number}{4} (\bibinfo{year}{2016}),
  \bibinfo{pages}{132:1--132:14}.
\newblock


\bibitem[\protect\citeauthoryear{Fisher, Ritchie, Savva, Funkhouser, and
  Hanrahan}{Fisher et~al\mbox{.}}{2012}]%
        {fisher12}
\bibfield{author}{\bibinfo{person}{Matthew Fisher}, \bibinfo{person}{Daniel
  Ritchie}, \bibinfo{person}{Manolis Savva}, \bibinfo{person}{Thomas
  Funkhouser}, {and} \bibinfo{person}{Pat Hanrahan}.}
  \bibinfo{year}{2012}\natexlab{}.
\newblock \showarticletitle{Example-based synthesis of {3D} object
  arrangements}.
\newblock \bibinfo{journal}{\emph{ACM Trans. on Graphics (Proc. SIGGRAPH
  Asia)}} \bibinfo{volume}{31}, \bibinfo{number}{6} (\bibinfo{year}{2012}),
  \bibinfo{pages}{135:1--11}.
\newblock


\bibitem[\protect\citeauthoryear{Fisher, Savva, Li, Hanrahan, and
  Nie{\ss}ner}{Fisher et~al\mbox{.}}{2015}]%
        {fisher15}
\bibfield{author}{\bibinfo{person}{Matthew Fisher}, \bibinfo{person}{Manolis
  Savva}, \bibinfo{person}{Yangyan Li}, \bibinfo{person}{Pat Hanrahan}, {and}
  \bibinfo{person}{Matthias Nie{\ss}ner}.} \bibinfo{year}{2015}\natexlab{}.
\newblock \showarticletitle{Activity-centric Scene Synthesis for Functional
  {3D} Scene Modeling}.
\newblock \bibinfo{journal}{\emph{ACM Trans. on Graphics (Proc. SIGGRAPH
  Asia)}} \bibinfo{volume}{34}, \bibinfo{number}{6} (\bibinfo{year}{2015}),
  \bibinfo{pages}{179:1--13}.
\newblock


\bibitem[\protect\citeauthoryear{Girshick}{Girshick}{2015}]%
        {girshick15}
\bibfield{author}{\bibinfo{person}{Ross Girshick}.}
  \bibinfo{year}{2015}\natexlab{}.
\newblock \showarticletitle{Fast {R-CNN}}. In \bibinfo{booktitle}{\emph{Proc.
  Int. Conf. on Computer Vision}}. \bibinfo{pages}{1440--1448}.
\newblock


\bibitem[\protect\citeauthoryear{Grover, Zweig, and Ermon}{Grover
  et~al\mbox{.}}{2019}]%
        {grover18}
\bibfield{author}{\bibinfo{person}{Aditya Grover}, \bibinfo{person}{Aaron
  Zweig}, {and} \bibinfo{person}{Stefano Ermon}.}
  \bibinfo{year}{2019}\natexlab{}.
\newblock \showarticletitle{Graphite: Iterative Generative Modeling of Graphs}.
  In \bibinfo{booktitle}{\emph{Proc. Conf. on Machine Learning}}.
\newblock


\bibitem[\protect\citeauthoryear{Hendrikx, Meijer, Van Der~Velden, and
  Iosup}{Hendrikx et~al\mbox{.}}{2013}]%
        {hendrikx13}
\bibfield{author}{\bibinfo{person}{Mark Hendrikx}, \bibinfo{person}{Sebastiaan
  Meijer}, \bibinfo{person}{Joeri Van Der~Velden}, {and}
  \bibinfo{person}{Alexandru Iosup}.} \bibinfo{year}{2013}\natexlab{}.
\newblock \showarticletitle{Procedural Content Generation for Games: A Survey}.
\newblock \bibinfo{journal}{\emph{ACM Trans. Multimedia Comput. Commun. Appl.}}
  \bibinfo{volume}{9}, \bibinfo{number}{1} (\bibinfo{year}{2013}),
  \bibinfo{pages}{1:1--1:22}.
\newblock


\bibitem[\protect\citeauthoryear{Johnson, Gupta, and Fei-Fei}{Johnson
  et~al\mbox{.}}{2018}]%
        {johnson18}
\bibfield{author}{\bibinfo{person}{Justin Johnson}, \bibinfo{person}{Agrim
  Gupta}, {and} \bibinfo{person}{Li Fei-Fei}.} \bibinfo{year}{2018}\natexlab{}.
\newblock \showarticletitle{Image Generation from Scene Graphs}. In
  \bibinfo{booktitle}{\emph{Proc. IEEE Conf. on Computer Vision \& Pattern
  Recognition}}.
\newblock


\bibitem[\protect\citeauthoryear{Li, Yang, Hertzmann, Zhang, and Xu}{Li
  et~al\mbox{.}}{2019b}]%
        {li19}
\bibfield{author}{\bibinfo{person}{Jianan Li}, \bibinfo{person}{Jimei Yang},
  \bibinfo{person}{Aaron Hertzmann}, \bibinfo{person}{Jianming Zhang}, {and}
  \bibinfo{person}{Tingfa Xu}.} \bibinfo{year}{2019}\natexlab{b}.
\newblock \showarticletitle{{LayoutGAN:} Generating Graphic Layouts with
  Wireframe Discriminators}. In \bibinfo{booktitle}{\emph{Proc. Int. Conf. on
  Learning Representations (ICLR)}}.
\newblock


\bibitem[\protect\citeauthoryear{Li, Patil, Xu, Chaudhuri, Khan, Shamir, Tu,
  Chen, Or, and Zhang}{Li et~al\mbox{.}}{2019a}]%
        {li2019_GRAINS}
\bibfield{author}{\bibinfo{person}{Manyi Li}, \bibinfo{person}{Akshay~Gadi
  Patil}, \bibinfo{person}{Kai Xu}, \bibinfo{person}{Siddhartha Chaudhuri},
  \bibinfo{person}{Owais Khan}, \bibinfo{person}{Ariel Shamir},
  \bibinfo{person}{Changhe Tu}, \bibinfo{person}{Baoquan Chen},
  \bibinfo{person}{Daniel~Cohen Or}, {and} \bibinfo{person}{Hao Zhang}.}
  \bibinfo{year}{2019}\natexlab{a}.
\newblock \showarticletitle{{GRAINS}: Generative Recursive Autoencoders for
  INdoor Scenes}.
\newblock \bibinfo{journal}{\emph{ACM Trans. on Graphics}}
  \bibinfo{volume}{38} (\bibinfo{year}{2019}).
\newblock


\bibitem[\protect\citeauthoryear{Liu, Wu, and Furukawa}{Liu
  et~al\mbox{.}}{2018}]%
        {liu18}
\bibfield{author}{\bibinfo{person}{Chen Liu}, \bibinfo{person}{Jiaye Wu}, {and}
  \bibinfo{person}{Yasutaka Furukawa}.} \bibinfo{year}{2018}\natexlab{}.
\newblock \showarticletitle{{FloorNet:} A Unified Framework for Floorplan
  Reconstruction from {3D} Scans}. In \bibinfo{booktitle}{\emph{Proc. Euro.
  Conf. on Computer Vision}}. \bibinfo{pages}{203--219}.
\newblock


\bibitem[\protect\citeauthoryear{Liu, Wu, Kohli, and Furukawa}{Liu
  et~al\mbox{.}}{2017}]%
        {liu17}
\bibfield{author}{\bibinfo{person}{C. Liu}, \bibinfo{person}{J. Wu},
  \bibinfo{person}{P. Kohli}, {and} \bibinfo{person}{Y. Furukawa}.}
  \bibinfo{year}{2017}\natexlab{}.
\newblock \showarticletitle{Raster-to-Vector: Revisiting Floorplan
  Transformation}. In \bibinfo{booktitle}{\emph{Proc. Int. Conf. on Computer
  Vision}}. \bibinfo{pages}{2214--2222}.
\newblock


\bibitem[\protect\citeauthoryear{Ma, Vining, Lefebvre, and Sheffer}{Ma
  et~al\mbox{.}}{2014}]%
        {ma14}
\bibfield{author}{\bibinfo{person}{Chongyang Ma}, \bibinfo{person}{Nicholas
  Vining}, \bibinfo{person}{Sylvain Lefebvre}, {and} \bibinfo{person}{Alla
  Sheffer}.} \bibinfo{year}{2014}\natexlab{}.
\newblock \showarticletitle{Game level layout from design specification}.
\newblock \bibinfo{journal}{\emph{Computer Graphics Forum}}
  \bibinfo{volume}{33}, \bibinfo{number}{2} (\bibinfo{year}{2014}),
  \bibinfo{pages}{95--104}.
\newblock


\bibitem[\protect\citeauthoryear{Merrell, Schkufza, and Koltun}{Merrell
  et~al\mbox{.}}{2010}]%
        {merrell10}
\bibfield{author}{\bibinfo{person}{Paul Merrell}, \bibinfo{person}{Eric
  Schkufza}, {and} \bibinfo{person}{Vladlen Koltun}.}
  \bibinfo{year}{2010}\natexlab{}.
\newblock \showarticletitle{Computer-generated Residential Building Layouts}.
\newblock \bibinfo{journal}{\emph{ACM Trans. on Graphics (Proc. SIGGRAPH
  Asia)}} \bibinfo{volume}{29}, \bibinfo{number}{6} (\bibinfo{year}{2010}),
  \bibinfo{pages}{181:1--181:12}.
\newblock


\bibitem[\protect\citeauthoryear{Merrell, Schkufza, Li, Agrawala, and
  Koltun}{Merrell et~al\mbox{.}}{2011}]%
        {merrell11}
\bibfield{author}{\bibinfo{person}{Paul Merrell}, \bibinfo{person}{Eric
  Schkufza}, \bibinfo{person}{Zeyang Li}, \bibinfo{person}{Maneesh Agrawala},
  {and} \bibinfo{person}{Vladlen Koltun}.} \bibinfo{year}{2011}\natexlab{}.
\newblock \showarticletitle{Interactive furniture layout using interior design
  guidelines}.
\newblock \bibinfo{journal}{\emph{ACM Trans. on Graphics (Proc. SIGGRAPH)}}
  \bibinfo{volume}{30}, \bibinfo{number}{4} (\bibinfo{year}{2011}),
  \bibinfo{pages}{87:1--10}.
\newblock


\bibitem[\protect\citeauthoryear{Rodrigues, Gaspar, and \'{A}lvaro
  Gomes}{Rodrigues et~al\mbox{.}}{2013a}]%
        {rodrigues13part1}
\bibfield{author}{\bibinfo{person}{Eug\'{e}nio Rodrigues},
  \bibinfo{person}{Ad\'{e}lio~Rodrigues Gaspar}, {and}
  \bibinfo{person}{\'{A}lvaro Gomes}.} \bibinfo{year}{2013}\natexlab{a}.
\newblock \showarticletitle{An evolutionary strategy enhanced with a local
  search technique for the space allocation problem in architecture, Part 1:
  Methodology}.
\newblock \bibinfo{journal}{\emph{Computer-Aided Design}} \bibinfo{volume}{45},
  \bibinfo{number}{5} (\bibinfo{year}{2013}), \bibinfo{pages}{887--897}.
\newblock


\bibitem[\protect\citeauthoryear{Rodrigues, Gaspar, and \'{A}lvaro
  Gomes}{Rodrigues et~al\mbox{.}}{2013b}]%
        {rodrigues13part2}
\bibfield{author}{\bibinfo{person}{Eug\'{e}nio Rodrigues},
  \bibinfo{person}{Ad\'{e}lio~Rodrigues Gaspar}, {and}
  \bibinfo{person}{\'{A}lvaro Gomes}.} \bibinfo{year}{2013}\natexlab{b}.
\newblock \showarticletitle{An evolutionary strategy enhanced with a local
  search technique for the space allocation problem in architecture, Part 2:
  Validation and performance tests}.
\newblock \bibinfo{journal}{\emph{Computer-Aided Design}} \bibinfo{volume}{45},
  \bibinfo{number}{5} (\bibinfo{year}{2013}), \bibinfo{pages}{898--910}.
\newblock


\bibitem[\protect\citeauthoryear{Rosser, Smith, and Morley}{Rosser
  et~al\mbox{.}}{2017}]%
        {rosser17}
\bibfield{author}{\bibinfo{person}{Julian~F. Rosser}, \bibinfo{person}{Gavin
  Smith}, {and} \bibinfo{person}{Jeremy~G. Morley}.}
  \bibinfo{year}{2017}\natexlab{}.
\newblock \showarticletitle{Data-driven estimation of building interior plans}.
\newblock \bibinfo{journal}{\emph{Int. J. of Geographical Information Science}}
  \bibinfo{volume}{31}, \bibinfo{number}{8} (\bibinfo{year}{2017}),
  \bibinfo{pages}{1652--1674}.
\newblock


\bibitem[\protect\citeauthoryear{Scarselli, Gori, Tsoi, Hagenbuchner, and
  Monfardini}{Scarselli et~al\mbox{.}}{2009}]%
        {scarselli09}
\bibfield{author}{\bibinfo{person}{F. Scarselli}, \bibinfo{person}{M. Gori},
  \bibinfo{person}{A.~C. Tsoi}, \bibinfo{person}{M. Hagenbuchner}, {and}
  \bibinfo{person}{G. Monfardini}.} \bibinfo{year}{2009}\natexlab{}.
\newblock \showarticletitle{The Graph Neural Network Model}.
\newblock \bibinfo{journal}{\emph{IEEE Trans. on Neural Networks}}
  \bibinfo{volume}{20}, \bibinfo{number}{1} (\bibinfo{year}{2009}),
  \bibinfo{pages}{61--80}.
\newblock


\bibitem[\protect\citeauthoryear{Simonovsky and Komodakis}{Simonovsky and
  Komodakis}{2018}]%
        {simonovsky18}
\bibfield{author}{\bibinfo{person}{Martin Simonovsky} {and}
  \bibinfo{person}{Nikos Komodakis}.} \bibinfo{year}{2018}\natexlab{}.
\newblock \showarticletitle{{GraphVAE:} Towards Generation of Small Graphs
  Using Variational Autoencoders}. In \bibinfo{booktitle}{\emph{Int. Conf. on
  Artificial Neural Networks}}. \bibinfo{pages}{412--422}.
\newblock


\bibitem[\protect\citeauthoryear{Sun, Zou, Tong, and Liu}{Sun
  et~al\mbox{.}}{2019}]%
        {sun19}
\bibfield{author}{\bibinfo{person}{Chun-Yu Sun}, \bibinfo{person}{Qian-Fang
  Zou}, \bibinfo{person}{Xin Tong}, {and} \bibinfo{person}{Yang Liu}.}
  \bibinfo{year}{2019}\natexlab{}.
\newblock \showarticletitle{Learning adaptive hierarchical cuboid abstractions
  of {3D} shape collections}.
\newblock \bibinfo{journal}{\emph{ACM Trans. on Graphics}}
  \bibinfo{volume}{38}, \bibinfo{number}{6} (\bibinfo{year}{2019}),
  \bibinfo{pages}{241:1--241:13}.
\newblock


\bibitem[\protect\citeauthoryear{Wang, Lin, Weissmann, Savva, Chang, and
  Ritchie}{Wang et~al\mbox{.}}{2019}]%
        {wang19}
\bibfield{author}{\bibinfo{person}{Kai Wang}, \bibinfo{person}{Yu-An Lin},
  \bibinfo{person}{Ben Weissmann}, \bibinfo{person}{Manolis Savva},
  \bibinfo{person}{Angel~X. Chang}, {and} \bibinfo{person}{Daniel Ritchie}.}
  \bibinfo{year}{2019}\natexlab{}.
\newblock \showarticletitle{{PlanIT:} Planning and Instantiating Indoor Scenes
  with Relation Graph and Spatial Prior Networks}.
\newblock \bibinfo{journal}{\emph{ACM Trans. on Graphics (Proc. SIGGRAPH)}}
  \bibinfo{volume}{38}, \bibinfo{number}{4} (\bibinfo{year}{2019}),
  \bibinfo{pages}{132:1--132:15}.
\newblock


\bibitem[\protect\citeauthoryear{Wu, Fan, Liu, and Wonka}{Wu
  et~al\mbox{.}}{2018}]%
        {wu18}
\bibfield{author}{\bibinfo{person}{Wenming Wu}, \bibinfo{person}{Lubin Fan},
  \bibinfo{person}{Ligang Liu}, {and} \bibinfo{person}{Peter Wonka}.}
  \bibinfo{year}{2018}\natexlab{}.
\newblock \showarticletitle{{MIQP-based} Layout Design for Building Interiors}.
\newblock \bibinfo{journal}{\emph{Computer Graphics Forum}}
  \bibinfo{volume}{37}, \bibinfo{number}{2} (\bibinfo{year}{2018}),
  \bibinfo{pages}{511--521}.
\newblock


\bibitem[\protect\citeauthoryear{Wu, Fu, Tang, Wang, Qi, and Liu}{Wu
  et~al\mbox{.}}{2019}]%
        {wu19}
\bibfield{author}{\bibinfo{person}{Wenming Wu}, \bibinfo{person}{Xiao-Ming Fu},
  \bibinfo{person}{Rui Tang}, \bibinfo{person}{Yuhan Wang},
  \bibinfo{person}{Yu-Hao Qi}, {and} \bibinfo{person}{Ligang Liu}.}
  \bibinfo{year}{2019}\natexlab{}.
\newblock \showarticletitle{Data-driven Interior Plan Generation for
  Residential Buildings}.
\newblock \bibinfo{journal}{\emph{ACM Trans. on Graphics (Proc. SIGGRAPH
  Asia)}} \bibinfo{volume}{38}, \bibinfo{number}{6} (\bibinfo{year}{2019}),
  \bibinfo{pages}{234:1--234:12}.
\newblock


\bibitem[\protect\citeauthoryear{Xu, Stewart, and Fiume}{Xu
  et~al\mbox{.}}{2002}]%
        {xu02}
\bibfield{author}{\bibinfo{person}{Ken Xu}, \bibinfo{person}{James Stewart},
  {and} \bibinfo{person}{Eugene Fiume}.} \bibinfo{year}{2002}\natexlab{}.
\newblock \showarticletitle{Constraint-based automatic placement for scene
  composition}. In \bibinfo{booktitle}{\emph{Proc. Graphics Interface}},
  Vol.~\bibinfo{volume}{2}. \bibinfo{pages}{25--34}.
\newblock


\bibitem[\protect\citeauthoryear{Yang, Wang, Vouga, and Wonka}{Yang
  et~al\mbox{.}}{2013}]%
        {yang13}
\bibfield{author}{\bibinfo{person}{Yong-Liang Yang}, \bibinfo{person}{Jun
  Wang}, \bibinfo{person}{Etienne Vouga}, {and} \bibinfo{person}{Peter Wonka}.}
  \bibinfo{year}{2013}\natexlab{}.
\newblock \showarticletitle{Urban Pattern: Layout Design by Hierarchical Domain
  Splitting}.
\newblock \bibinfo{journal}{\emph{ACM Trans. on Graphics (Proc. SIGGRAPH
  Asia)}} \bibinfo{volume}{32}, \bibinfo{number}{6} (\bibinfo{year}{2013}),
  \bibinfo{pages}{181:1--181:12}.
\newblock


\bibitem[\protect\citeauthoryear{You, Ying, Ren, Hamilton, and Leskovec}{You
  et~al\mbox{.}}{2018}]%
        {you18}
\bibfield{author}{\bibinfo{person}{Jiaxuan You}, \bibinfo{person}{Rex Ying},
  \bibinfo{person}{Xiang Ren}, \bibinfo{person}{William~L. Hamilton}, {and}
  \bibinfo{person}{Jure Leskovec}.} \bibinfo{year}{2018}\natexlab{}.
\newblock \showarticletitle{{GraphRNN:} Generating Realistic Graphs with Deep
  Auto-regressive Model}. In \bibinfo{booktitle}{\emph{Proc. Conf. on Machine
  Learning}}.
\newblock


\bibitem[\protect\citeauthoryear{Zhang, Yang, Ma, Luo, Huth, Vouga, and
  Huang}{Zhang et~al\mbox{.}}{2018}]%
        {zhang2019hybrid}
\bibfield{author}{\bibinfo{person}{Zaiwei Zhang}, \bibinfo{person}{Zhenpei
  Yang}, \bibinfo{person}{Chongyang Ma}, \bibinfo{person}{Linjie Luo},
  \bibinfo{person}{Alexander Huth}, \bibinfo{person}{Etienne Vouga}, {and}
  \bibinfo{person}{Qixing Huang}.} \bibinfo{year}{2018}\natexlab{}.
\newblock \showarticletitle{Deep Generative Modeling for Scene Synthesis via
  Hybrid Representations}.
\newblock \bibinfo{journal}{\emph{CoRR}}  \bibinfo{volume}{abs/1808.02084}
  (\bibinfo{year}{2018}).
\newblock
\showeprint[arxiv]{1808.02084}
\urldef\tempurl%
\url{http://arxiv.org/abs/1808.02084}
\showURL{%
\tempurl}


\bibitem[\protect\citeauthoryear{Zhao, Hu, Guerrero, Mitra, and Komura}{Zhao
  et~al\mbox{.}}{2016}]%
        {zhao16}
\bibfield{author}{\bibinfo{person}{Xi Zhao}, \bibinfo{person}{Ruizhen Hu},
  \bibinfo{person}{Paul Guerrero}, \bibinfo{person}{Niloy Mitra}, {and}
  \bibinfo{person}{Taku Komura}.} \bibinfo{year}{2016}\natexlab{}.
\newblock \showarticletitle{Relationship Templates for Creating Scene
  Variations}.
\newblock \bibinfo{journal}{\emph{ACM Trans. on Graphics (Proc. SIGGRAPH
  Asia)}} \bibinfo{volume}{35}, \bibinfo{number}{6} (\bibinfo{year}{2016}),
  \bibinfo{pages}{207:1--13}.
\newblock


\end{thebibliography}
